\documentclass[runningheads]{llncs}

\usepackage{eccv}

\usepackage{eccvabbrv}

\usepackage{graphicx}
\usepackage{booktabs}

\usepackage[accsupp]{axessibility}  

\usepackage{hyperref}

\usepackage{orcidlink}

\usepackage{enumitem}
\usepackage{makecell}
\usepackage{multirow}
\usepackage{adjustbox}

\definecolor{mygreen}{RGB}{0,128,0}

\definecolor{myred}{RGB}{220,20,60}
\newcommand{\R}[1]{\textcolor{myred}{$(#1)$}}
\newcommand{\G}[1]{\textcolor{mygreen}{$(#1)$}}
\newcommand{\na}{\textcolor{gray}{$n/a$}}

\newcommand{\myPara}[1]{\vspace{.05in}\noindent\textbf{#1.}}

\usepackage{tikz}

\usepackage{setspace}

\usepackage{xspace}
\newcommand{\promptname}{$\mathcal{API}$\xspace}

\makeatletter
\renewcommand
\tableofcontents
{
  \vspace{2em}
  \begin{center}
  \begingroup
  \Large
  \textbf{Contents}
  \endgroup
  \end{center}
  \vspace{-1em}
  \begin{spacing}{1.75}
  \large
  {
  \@starttoc{toc}%
  }
  \end{spacing}
}
\makeatother

\usepackage{fontawesome}
\usepackage{xspace}
\newcommand{\huggingface}{\raisebox{-1.5pt}{\includegraphics[height=1.05em]{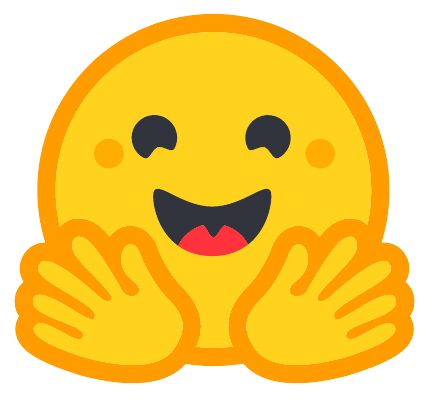}}\xspace}
\newcommand{\github}{\raisebox{-1.5pt}{\includegraphics[height=1.05em]{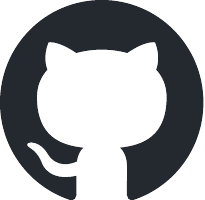}}\xspace}

\newcommand{\worldwideweb}{\raisebox{-1.5pt}{\includegraphics[height=1.05em]{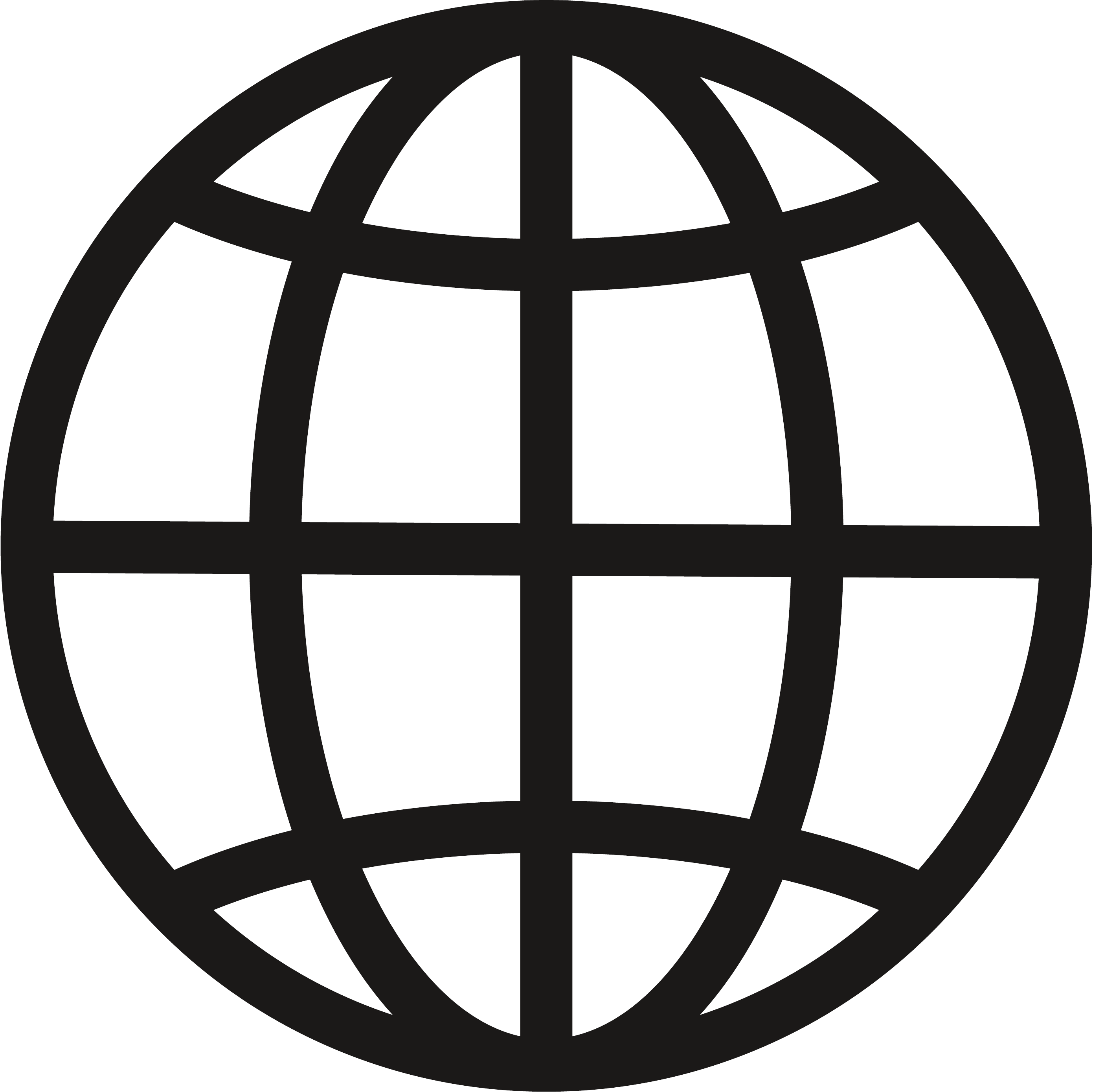}}\xspace}

\begin{document}

\title{Attention Prompting on Image for Large Vision-Language Models} 

\titlerunning{Attention Prompting on Image for Large Vision-Language Models}

\author{Runpeng Yu\orcidlink{0000-0001-6321-9614} \and
Weihao Yu$^\dagger$\orcidlink{0000-0003-3349-5890} \and
Xinchao Wang$^\dagger$\orcidlink{0000-0003-0057-1404}}

\authorrunning{Yu et al.}

\institute{National University of Singapore\\
\email{\{r.yu,weihaoyu\}@u.nus.edu}, \email{xinchao@nus.edu.sg}}
\renewcommand{\thefootnote}{}
\footnotetext{$^\dag$ Corresponding author.}
\renewcommand{\thefootnote}{\arabic{footnote}}

\maketitle

\begin{center}
    \renewcommand{\arraystretch}{1.2}
    \begin{tabular}{rll}
        \worldwideweb & \textbf{Website} & \url{https://yu-rp.github.io/api-prompting/}\\
        \github & \textbf{Code} & \url{https://github.com/yu-rp/apiprompting}\\
        \huggingface & \textbf{Space} & \url{https://huggingface.co/spaces/rp-yu/apiprompting}\\
    \end{tabular}
\end{center}
\begin{figure}[!h]
  \centering
  \includegraphics[width=0.88\textwidth]{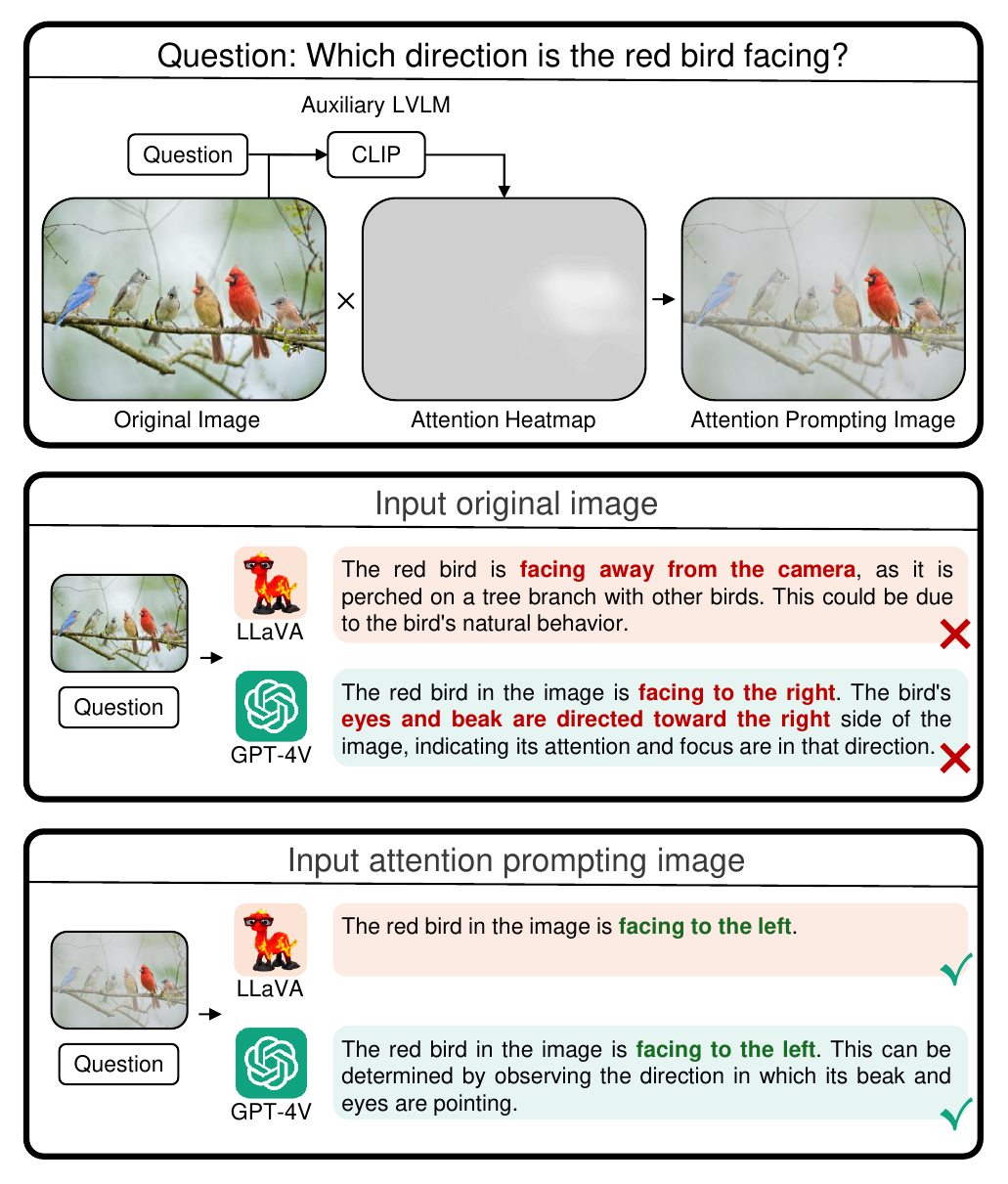}
  \caption{
  Comparison of the proposed Attention Prompting on Image (\promptname) with the naive VQA.\promptname provides hints for LVLM by simply overlying a heatmap on the image. 
  }
  \label{fig:head}
\end{figure}
\clearpage
\begin{abstract}
    Compared with Large Language Models (LLMs), Large Vision-Language Models (LVLMs) can also accept images as input, thus showcasing more interesting emergent capabilities and demonstrating impressive performance on various vision-language tasks. Motivated by text prompting in LLMs, visual prompting has been explored to enhance LVLMs' capabilities of perceiving visual information. However, previous visual prompting techniques solely process visual inputs without considering text queries, limiting the models' ability to follow text instructions to complete tasks. To fill this gap, in this work, we propose a new prompting technique named Attention Prompting on Image (\promptname), which just simply overlays a text-query-guided attention heatmap on the original input image and effectively enhances LVLM on various tasks. Specifically, we generate an attention heatmap for the input image dependent on the text query with an auxiliary model like CLIP. Then the heatmap simply multiplies the pixel values of the original image to obtain the actual input image for the LVLM. Extensive experiments on various vison-language benchmarks verify the effectiveness of our technique. For example, \promptname improves LLaVA-1.5 by 3.8\% and 2.9\% on MM-Vet and LLaVA-Wild benchmarks, respectively.
    
\end{abstract}
\section{Introduction}
Benefiting from the great progress of Large Language Models (LLMs) \cite{touvron2023llama, touvron2023llama2, achiam2023gpt}, Large Vision-Language Models (LVLMs) \cite{alayrac2022flamingo,awadalla2023openflamingo,gpt4v,llava,zhu2023minigpt,yang2023mm,gao2023llama,li2023otter,dai2023instructblip} also advances rapidly, represented by the seminal works GPT-4V \cite{gpt4v} and LLaVA \cite{llava}.\footnote{Although also referred to as Multimodal Large Language Model (MLLM) or Large Multimodal Model (LMM)~\cite{llava, gpt4v}, we use Large Vision-Language Model (LVLM) to refer to the models discussed in this paper, as we primarily utilizes the model's vision and language capabilities.} They have been widely applied in tasks that involve understanding both visual and linguistic information, such as referring segmentation~\cite{vlm_app_2,vlm_app_1}, localization~\cite{vlm_app_1}, captioning~\cite{vlm_app_6}, open world 2D/3D understanding~\cite{vlm_app_3,gpt4v,gemini,vlm_app_4}, and image editing~\cite{gpt4v,vlm_app_5}. 

To enhance the performance of LVLMs, an economical method is to develop prompting techniques to elicit the models' potential. Similar to textual prompting \cite{wei2022chain,kojima2022large}, visual prompting\footnote{In this work, we specifically use ``visual prompts'' to refer to masks, circles, marks, and other annotations added to images and use ``visual prompting'' to refer to technologies that employ visual prompts to assist in VQA tasks.} \cite{fgvp,som} is a technique that enhances a model's understanding of images by directly adding annotations such as masks, circles, and marks to the image. This technique provides clear hints for visual perception by highlighting areas relevant to solving the problem, guiding the model's attention to specific parts of the image, thus mitigating issues arising from complex scenes with distractions. It has been demonstrated that even simple visual cues like circles\cite{vp_clip_1}, arrows~\cite{gpt4v}, or image tiling~\cite{SPHINX} can improve LVLMs' ability to extract the required information correctly. Unlike methods that improve LVLM performance through adaptation or fine-tuning, visual prompting does not require the training process, thereby reducing the risks of overfitting and knowledge forgetting. Moreover, compared to textual prompts, visual prompting is more direct and precise in guiding the model's focus to specific areas. Textual descriptions cannot succinctly describe an irregular area in an image or accurately indicate the location of a specific object, and they also face issues with aligning textual coordinates with actual image pixels \cite{gpt4v}. However, compared to research on textual prompts and LVLM fine-tuning, visual prompting is still underexplored. 

Previous visual prompting techniques focused on designing appropriate fine-grained annotations to the image, aiming to highlight important local areas without impairing the model's overall understanding of the image. Remarkably, FGVP \cite{fgvp} and SoM \cite{som} are both based on segmentation masks \cite{som}: The former blurs the image outside the segmentation mask while the latter overlays the image with a set of including alphanumerics, masks, and boxes. However, all these methods sorely process the input images without considering the text query content. In other words, whatever the text query is, an image's visual prompting results are the same. This can easily lead to a mismatch between the prompted image and the text query, as different text queries for the same image require focus on different areas and necessitate different annotations. This mismatch may thereby limit the model's ability to follow instructions accurately.

To address this issue, in this paper, we propose a novel prompting technique named Attention Prompting on Image (\promptname), which just simply overlays a text-query-guided attention heatmap on the original input image. Specifically, to generate text-guided attention heatmap for an image, we utilize an auxiliary LVLM that can accept both image and text as input. For image-text matching type (like CLIP \cite{clip}) as auxiliary model, we devised a heatmap generation technique based on the decomposition of cls token similarity score. For the vision-language-input text generation model (like LLaVA \cite{llava}), we generate the heatmap based on attention weights. Extensive experiments on various commonly used vision-language (VL) datasets verify the effectiveness of \promptname in enhancing the VLM's perception of visual information. 
For example, \promptname improves LLaVA-1.5 by 3.8\%, 2.9\%, and 2.3\% on MM-Vet, LLaVA-Bench and MMMU benchmarks

Our contributions can be summarized as follows:
\begin{enumerate}
    \item We find that current visual prompting techniques sorely modify input images without considering the text query, limiting the model's capability to follow instructions accurately.
    \item To fill the gap, we propose the \promptname method, exploring how to derive valuable attribution maps from various types of VLM models and utilize them as visual prompts to offer hints for visual perception, thereby boosting performance.
    \item Our experiments demonstrate the effectiveness of our method across a wide range of VLM models on various datasets. Moreover, our approach has also proven effective in addressing the issue of hallucination.
\end{enumerate}

\section{Related Works}
\subsection{Visual Prompting for LVLM}

Originating from language models~\cite{language_prompt_1,language_prompt_2,ma2023llm}, the concept of prompting has been widely applied in vision models and vision language models to enhance the transfer learning and adaptation for various tasks (\textit{e.g., }classification~\cite{vlm_prompt_classification,coop,cocoop,vpt, vlpt}, detection~\cite{vlm_prompt_detection,prompt_detection,prompt_open_vocab_detection}, segmentation~\cite{vlm_prompt_segmentation} and generation~\cite{prompt_gan}) and under various learning settings (\textit{e.g.,} few-shot learning~\cite{prompt_one_shot}, continual learning~\cite{prompt_cl_1,prompt_cl_2,prompt_cl_3}, domain adaptation/generalization~\cite{prompt_da,prompt_dg}, unlearning~\cite{prompt_unlearning}, and long-tailed learning~\cite{prompt_long_tail}). It is crucial to distinguish our work from soft prompts generated through gradient optimization and related prompt-tuning efforts. These prompts, concatenated in the form of continuous vectors to the token sequence of the VL model's transformer layer input~\cite{vlm_soft_prompt_1,vlm_soft_prompt_2,vlm_soft_prompt_3}, or added to the input image as optimizable pixel patches and paddings~\cite{vlm_pixel_prompt_1,vlm_pixel_prompt_2}, depend on an additional learning process. Thus, they are strongly coupled with the model and dataset, lack generalizability, and are not intuitively interpretable. Moreover, since (part of) these prompts are incorporated at a shallow layer, their optimization process involves gradient propagation throughout the entire branch, which is costly. Unlike these methods, the visual prompting studied in this paper is manually designed and automatically generated by extra LVLMs. It is interpretable and generalizable across different models and tasks.

Visual prompting is a specialized technique in vision models, especially for segmentation tasks~\cite{sam,vp_3,vp_2}. Based on an additionally trained prompt encoder, manually annotated points, strokes, boxes, and irregular masks can provide these models with extra instructions to assist in controlling segmentation granularity or in facilitating instance selection. Recently, LVLMs have also been shown to understand manually added circles and color masks in images in a zero-shot manner, focusing attention on highlighted areas without relying on additional encoder components~\cite{vp_clip_1,vp_clip_2}. Unlike these works that explore the LVLM's ability to understand visual prompts, our method discusses how to use pretrained LVLMs to automatically generate visual prompts to enhance image readability. 

The two methods most related to ours are \cite{som} and \cite{fgvp}, which modify masks generated by segmentation models to construct visual prompts to improve LVLM's performance in segmentation and grounding tasks. Our method differs fundamentally from theirs in that we use LVLMs to construct visual prompts. This leads to two main differences in functionality and applicability. 1) For a single image, the visual prompts generated by \cite{som,fgvp} are invariant, as these models rely on fixed segmentation models. In contrast, with different text queries, our method can adapt and generate distinct visual prompts to emphasize different areas as required. 2) The visual prompts generated by \cite{som,fgvp} are essentially instance-specific proposals for segmentation and grounding tasks, focusing on enhancing the LVLM's grounding capability. Conversely, our visual prompts aim to highlight important areas needed to address text queries, thereby improving the LVLM's performance in general Visual Question Answering tasks.

\subsection{Self-Reflection and Ensemble}
Our method involves LVLM at two stages: once for generating visual prompts and once for performing inference. When the same LVLM is used at both stages, our approach can be seen as a method to enhance LVLM performance using self-reflection technology. The concept of Self-Reflection originated from LLMs \cite{pan2023automatically,shinn2024reflexion} but can be directly transferred to LVLMs. Self-Reflection scheme improves model performance by repeatedly answering a query and iteratively updating the answer. The Self-Reflection process involves using self-evaluation~\cite{sr_6}, self-checking~\cite{sr_5}, self-feedback~\cite{sr_4}, feedback from the external environment~\cite{sr_2,sr_3}, and even previous answers themselves~\cite{sr_1} as hints to input into the model for it to answer the question again. Unlike these works, where the medium of self-reflection is text, our method employs visual prompting to achieve Self-Reflection in the pixel space.

When different LVLMs are used at two stages, our method can be considered a form of model ensemble, where the knowledge of the first VLM is ensembled into the second VLM in the form of visual prompts. In tasks with standard outputs, deep learning model ensemble involves aggregating outputs from multiple models~\cite{dl_ensemble}. However, output aggregation is invalid in generation tasks. In LLMs and LVLMs, model ensemble is achieved in the form of sequential or stage-wise use between auxiliary and inference models. The final inference model can enhance its performance by incorporating outputs from other auxiliary models into its input. The auxiliary model outputs used as inputs for the inference model can be responses from another language model~\cite{vlm_ensemble_1} or textualized outputs from vision models or vision-language models (image captions, category names)~\cite{vlm_ensemble_2,vlm_ensemble_3}. Unlike these works, our method uses visual rather than textual signals for ensembling. Furthermore, our approach does not ensemble the final hard outputs of auxiliary models but their visual cues used during the inference process. This soft knowledge ensemble provides valuable auxiliary information and mitigates error accumulation introduced by mistakes in auxiliary model inference.
\section{Method}
Large Vision Language Model $f$ takes an image $I \in \mathbb{R}^{H \times W \times 3}$ and a text query $T^i$ as inputs, generating an output text $T^o = f(I, T^i)$. During the inference process using \promptname, instead of being directly fed into $f$, the original image $I$ undergoes an additional annotation operation $\mathcal{A}$, resulting in an image $I^a = \mathcal{A}(I, T^i)$ that has been overlaid with a heatmap $\Phi$. Subsequently, the annotated image $I^a$ and the original query are input into the LVLM model $f$, producing the output $T^o = f(I^a, T^i)$. The overall framework of the method is shown in \cref{fig:head}.

In our method, the annotation process comprises two steps. The first step involves using an auxiliary LVLM model $g$ to establish an initial attribution map $\Psi$ between the text query $T^i$ and each patch of the image. This attribution map indicates which patches in the image are more relevant to $T^i$ or which patches should be paid more attention to for answering $T^i$. In our method, there are no additional constraints on the LVLM $g$; if the inference LVLM $f$ is accessible and capable of performing the annotation operation $\mathcal{A}$, then the LVLM $g$ used to generate the attribution map can be the same as $f$, \textit{i.e.}, $g = f$. Alternatively, $g$ could be a different LVLM to introduce knowledge from other models to enhance $f$'s functionality, \textit{i.e.},  $g \neq f$. Moreover, due to the diversity of LVLM models, we do not necessarily use the attention map as our attribution map. For example, for the image-text matching model, experiments have shown that using the attention map as the attribution map has suboptimal results. After obtaining the attribution map $\Psi$, the second step in the annotation process is to convert it into a suitable $\Phi$ and apply it to the original image using alpha blending.

Various LVLM models can be utilized to generate attribution maps. We discuss two prevalent and representative LVLM models: CLIP~\cite{clip}, exemplifying image-text matching models, and LLaVA~\cite{llava}, representing vision-language-input text generation models.

\subsection{Obtaining Attribution Map from CLIP}
The CLIP model, $g_{\text{clip}}$, consists of an image encoder and a text encoder, calculating the similarity between an image and a text query in the image-language latent space, $sim(\hat{I}, \hat{T})$, where $\hat{I} = g_{\text{clip}}^{\text{img}}(I)$ and $\hat{T} = g_{\text{clip}}^{\text{text}}(T)$. This similarity measure evaluates the correlation between the entire image and the query. To obtain the attribution map value from the text query to each image patch, we decompose the output image-level similarity $\hat{I}$ and then calculate the similarity of each patch's output with the $\hat{T}$. 

The decomposition process is as follows. Due to the presence of residual connections, the final output of the vision tower, $\hat{I}$, actually includes influences from each layer. Consequently, $\hat{I}$ can be expressed as a linear combination of the values at the class token positions from each layer
\begin{equation}
    \hat{I} =  \mathcal{L}(\left[Z^{0}_{\text{cls}}\right]) + \sum_{l=1}^{L} \mathcal{L}(\left[\text{MSA}^l(Z^{l-1})\right]_{\text{cls}}) + \sum_{l=1}^{L} \mathcal{L}( \left[\text{MLP}^l(\hat{Z}^l)\right]_{\text{cls}}),
\end{equation}
where $L$ denotes the number of transformer layers within the vision encoder, with MSA and MLP representing the Multihead Self-Attention structure and the Multi-Layer Perceptron structure within the transformer, respectively; $\mathcal{L}$ represents the linear transformation that includes the fully-connected layer and the normalization operations performed after the transformer structure, before calculating the similarity score; $Z^l$ signifies the input token sequence for the $l$-th transformer layer; and $[Z]_{\text{cls}}$ indicates the value of the cls token within the token sequence $Z$. These output cls tokens are aggregated through residual connections to form the output of the vision encoder. As evidenced in \cite{llava,clipprs}, among these summation terms, the outputs of the last few layers of MSA play a decisive role, while the contributions from the outputs of the shallow MSA layers, the outputs of MLP, and the $Z^0_{\text{cls}}$ term, which is independent of the input image, can be considered negligible to the final measurement of similarity. Therefore, the similarity $sim(\hat{I},\hat{T})$ can effectively be approximated by calculating the similarity between $\hat{T}$ and the aggregated outputs of MSAs in the deeper layers :
\begin{equation}\label{eq:sim_approx}
    sim(\hat{I},\hat{T}) \approx sim(\sum_{l=L'}^{L} \mathcal{L}(\left[\text{MSA}^l(Z^{l-1})\right]_{\text{cls}}) ,\hat{T}),
\end{equation}
where $L'$ represents a predefined starting layer index. To further calculate the attribution of the text query to each patch, inspired by \cite{clipprs}, we unfold the operations of the Multihead Self-Attention, obtaining
\begin{align}
    \left[\text{MSA}^l(Z^{l-1})\right]_{\text{cls}} 
    &= \sum_{h}^H \left[A^{(l,h)}V^{(l,h)}W^{(l,h)}\right]_{\text{cls}} + B^{l}\\
    &= \sum_{t=1}^T\underbrace{\left[\sum_{h}^H A^{(l,h)}_{\text{cls},t}V^{(l,h)}_{t,:}W^{(l,h)} + \frac{1}{HT}B^{l}\right]}_{\text{\parbox{5cm}{\centering The MSA output corresponding to \newline the $t$-th patch(token)}}} \triangleq \sum_{t=1}^T\eta^l_t,
\end{align}
where $A^{(l,h)}$, $V^{(l,h)}$ are the attention map and the value matrix in the $l$-th layer corresponding to the $h$-th head, respectively; $W^{(l,h)}$ is the weight matrix in the $l$-th layer used to merge the multiple attention heads and corresponds to the $h$-th head; $B^{(l)}$ is the bias matrix in the $l$-th layer used to merge the multiple attention heads; $A^{(l,h)}_{\text{cls},t}$ denotes the attention value of the class token towards the $t$-th token in $A^{(l,h)}$, and $V^{(l,h)}_{t,:}$ represents the $t$-th row of $V^{(l,h)}$; $H$ and $T$ are the number of attention heads and the number of tokens, respectively; and the value $T$ equals the number of patches $P\times P$ plus one. 

Consequently, by summing across layers and incorporating the final linear transformation, we obtain $\psi_t\triangleq\sum_{l=L'}^L\mathcal{L}(\eta^l_t)$, which is the direct influence of the $t$-th patch to the similarity in \cref{eq:sim_approx}, allowing us to calculate the similarity between text query and the $t$-th image patch. Accordingly, the attribution map $\Psi^{cls}\in\mathbb{R}^{P\times P}$ is defined as 
\begin{align}
    \Psi^{cls}_{i,j} &\triangleq sim(\psi_t,\hat{T}), \qquad\text{where}\  t=1 + j+P*(i-1).
\end{align}

By decomposing the cls token, we can identify which patches are more relevant to the query. This approach is particularly effective when the query contains specific entities, allowing for accurate grounding. However, in complex Visual Question Answering (VQA) tasks, there are often no explicit entities mentioned in the query, or the logic and analysis process involved in answering the question may rely on entities that are not explicitly mentioned in the query. To address this issue, we also define another complementary attribution map $\Psi^{comp}$ using the CLIP model. This map is designed to capture patches that have potential or implicit relevance to the query.

We experimentally observe that, in the vision transformer of CLIP, the similarity score of the query feature $\hat{T}$ and tokens other than the cls token in the final layer can (inversely) select the important regions. Patches corresponding to the image background or large monochrome areas have a significantly higher similarity score with $\hat{T}$ than those tokens representing specific entities (which may not necessarily appear in the query). This phenomenon is similar to observations made in \cite{vit_reg}. Drawing on analyses of the transformer's mechanism in \cite{vit_reg,transformer_mechanism}, a potential explanation is that these ``blank'' tokens, lacking valuable information themselves, are treated by the transformer as registers. The transformer initially utilizes them to store information from other informative tokens, subsequently filtering and aggregating this stored information to the class token via the attention mechanism to formulate the final prediction. Therefore, tokens other than the class token, with a high similarity score to $\hat{T}$, represent patches with low information content that can be disregarded. We define the complementary attribution map as follows
\begin{align}
    \Psi^{comp}_{i,j} &\triangleq 1 - sim(\mathcal{L}(Z^L_t),\hat{T}), \qquad\text{where}\  t=1 + j+P*(i-1),
\end{align}
where $Z^L_t$ is the $t$-th output token from the last transformer layer. The complementary attribution map is inversely related to similarity, suggesting that patches lacking information are ignored, retaining only those with potential relevance.

Thus, we obtain two attribution maps that complement each other: $\Psi^{cls}$ explicitly identifies patches directly related to entities in the query but may miss some potentially relevant patches. $\Psi^{comp}$ equally identifies all patches with potential relevance but lacks specificity and cannot highlight those directly related to entities in the query.

By integrating the two attribution maps through the following operation, we obtain the final attribution map for CLIP:
\begin{equation}
    \Psi_{i,j} \triangleq  \Psi^{cls}_{i,j} + \Psi^{comp}_{i,j} -  \Psi^{comp}_{i,j} * \Psi^{cls}_{i,j}.
\end{equation}
This integration can be considered as a soft OR operation (a detailed mathematical explanation is provided in the Appendix). This ensures that the final attribution map highlights patches directly related to entities within the query while retaining those with potential or implicit relevance, merely reducing the weights of patches that do not contain important information for the query. If the function of the final attribution map were described as an algorithm, then this attribution map would, in the first step, apply a mask to all non-informative patches, making them less considered in subsequent VQA processes while leaving other patches unaffected; and, in the second step, for patches not masked, if a patch is directly related to the entities in the query, it further highlights this patch.

\subsection{Obtaining Attribution Map from LLaVA}

The LLaVA model is an auto-regressive vision-language-input text generation model that utilizes Multihead Self-Attention to extract information from text queries and image patches, predicting the following tokens. Given a text token sequence of length $N$, $Z^{\text{text}} = \{Z_{t}^{\text{text}}\}_{t=1}^{N}$, and an image token sequence of length $P\times P$, $Z^{\text{img}} = \{Z_{t}^{\text{img}}\}_{t=1}^{P\times P}$, LLaVA generates a new token sequence of length $M$, $Z^{\text{out}} = \{Z_{t}^{\text{out}}\}_{t=1}^{M}$. We directly use the attention weight between token $Z_{t}^{\text{out}}$ and each image token as $Z_{t}^{\text{out}}$'s attribution to that image patch. Similar to the strategy for the CLIP model, we select attention maps from the deeper layer to extract attention weights. The final attribution map is averaged over the entire generated token sequence and all attention heads. Formally, the attribution map $\Psi$ is defined as

\begin{align}
    \Psi_{i,j} &\triangleq \frac{1}{MH}\sum_{m=1}^M\sum_{h=1}^H A^{(\bar{L},h)}_{m,t}, \qquad\text{where}\  t=j+P*(i-1).
\end{align}
In the definition, $A^{(\bar{L},h)}$ is again the attention map in the $\bar{L}$-th layer corresponding to the $h$-th head, where $\bar{L}$ is a set to be a hyper-parameter; for notation simplicity, $A^{(\bar{L},h)}$ here is a submatrix of the entire attention map and only includes cross attention between $Z^{\text{out}}$ and  $Z^{\text{img}}$;  $A^{(\bar{L},h)}_{m,t}$ still denotes the attention value from the $m$-th token to the $t$-th token. 

\subsection{From Token Space to Pixel Space}

The attribution map $\Psi\in\mathbb{R}^{P\times P}$ is generated in the token space. We first resize it back to the pixel space to obtain the raw heatmap $\hat{\Phi}\triangleq \text{Resize}(\Psi)$. Due to the square shape of the patches, the mask pattern in $\hat{\Phi}$ also appears rectangular. To mitigate the issue that the rectangular mask pattern does not align with the object's irregular shape, we apply a mean filter to obtain the final heatmap $\Phi\triangleq \text{Mean}_k(\hat{\Phi})$, where $k$ is the kernel size of the filter. The final heatmap $\Phi$ is then overlaid on the original image by using it as the alpha channel, resulting in the final image after annotation $I^a$.
\section{Experiments}
We show the main experimental results in this section. More experiments and implementation details are in the appendix.

\begin{table}[tb]
  \caption{Comparison of our method with previous textual and visual prompting methods for various LVLMs. The best result are marked for each model-dataset pair.
  }
  \label{tab:main}
\begin{adjustbox}{center, max width=\textwidth}
\scriptsize{
\begin{tabular}{@{}clcccccc@{}}
\toprule
\multirow{2}{*}{\makecell{Inference\\Model}} & \multirow{2}{*}{Prompting Method} & \multicolumn{6}{c}{Dataset} \\ \cmidrule(l){3-8} 
                                    &  & VisWiz         & TextVQA         & MMMU            & MM-Vet          & MME & LLaVA-Bench \\ \midrule
\multirow{7}{*}{LLaVA}              & w/o prompt                  & 60.93           & 48.32           & 35.15           & 32.8           & 85.5                        & 71.9      \\
                                    & $\quad+$Step-by-Step                      & 60.98 \G{+0.1}         & 48.22 \R{-0.1}         & 35.40 \G{+0.3}         & 33.7 \G{+0.9}         & 84.2 \R{-1.3}                      & 73.5 \G{+1.6}    \\
                                    & FGVP (Mask)                  & 56.89 \R{-4.0}         & 39.38 \R{$<-\ $5} & 36.14 \G{+1.0}         & 31.0 \R{-1.8}         & 75.8 \R{$<-\ $5}              & 57.4 \R{$<-\ $5} \\
                                    & FGVP (RBM)     & 61.22 \G{+0.3}         & 33.91 \R{$<-\ $5} & 35.00 \R{-0.2}         & 25.0 \R{$<-\ $5} & 81.4 \R{-4.1}                      & 57.4 \R{$<-\ $5}    \\
                                    & SoM & 54.16 \R{$<-\ $5} & 18.81 \R{$<-\ $5} & 35.57 \G{+0.4}         & 26.4 \R{$<-\ $5} & 75.4 \R{$<-\ $5}              & 56.1 \R{$<-\ $5}                     \\
                                    & Ours (CLIP)                  & 61.26 \G{+0.3}         & 48.78 \G{+0.5}         & \textbf{37.52} \G{+2.4}         & 35.3 \G{+2.5}          & \textbf{87.2} \G{+1.7}                      & 74.1 \G{+2.2}     \\
                                    & Ours (LLaVA)                 & \textbf{61.35} \G{+0.4}         & \textbf{48.79} \G{+0.5}         & 36.95 \G{+1.8}          & \textbf{36.6} \G{+3.8}          & 86.3 \G{+0.8}                      & \textbf{74.8} \G{+2.9}     \\\midrule
\multirow{7}{*}{CogVLM}             & w/o prompt                   & 53.54                  & 78.41                     & 36.43                 & 49.6                    & 81.8               & 50.8      \\
                                    & $\quad+$Step-by-Step                      & 28.86 \R{$<-\ $5} & 42.53 \R{$<-\ $5} & 29.19 \R{$<-\ $5} & 48.0 \R{-1.6}          & 63.0 \R{$<-\ $5}              & 40.7 \R{$<-\ $5}                     \\
                                    & FGVP (Mask)                  & 53.55 \G{+0.0}         & 63.69 \R{$<-\ $5} & 35.34 \R{-1.1}         & 44.1 \R{$<-\ $5} & 80.4 \R{-1.4}                      & 49.1 \R{-1.7}    \\
                                    & FGVP (RBM)     & 53.68 \G{+0.1}         & 65.51 \R{$<-\ $5} & 36.55 \G{+0.1}         & 48.2 \R{-1.4}         & 82.0 \G{+0.2}                      & 48.1 \R{-2.7}    \\
                                    & SoM & 51.00 \R{-2.5}          & 36.64 \R{$<-\ $5} & 35.55 \R{-0.9}         & 31.2 \R{$<-\ $5}  & 78.0 \R{-3.8}                       & 38.9 \R{$<-\ $5}                     \\
                                    & Ours (CLIP)                  & 54.01 \G{+0.5}         & \textbf{78.99} \G{+0.6}         & \textbf{37.05} \G{+0.6}         & \textbf{52.5} \G{+2.9}          & 82.3 \G{+0.5}                      & \textbf{53.3} \G{+2.5}    \\
                                    & Ours (LLaVA)                 & \textbf{54.34} \G{+0.8}         & 78.85 \G{+0.4}         & 36.95 \G{+0.5}         & 52.0 \G{+2.4}          & \textbf{82.7} \G{+0.9}                      & 52.4 \G{+1.6}    \\\midrule
\multirow{7}{*}{\makecell{GPT-4V\\(1106)}}                          & w/o prompt                   & 59.40                              & 50.60                                 & 50.55                                    	& 67.00                                                & 84.3                        & 102.0      \\
                                                                    & $\quad+$Step-by-Step         & 55.75 \R{-3.6}  					& 49.85 \R{-0.7}      					& 48.33 \R{-2.2}         					& 62.50 \R{-4.5}          		                      & 82.0 \R{-2.3}                      & 102.6 \G{+0.6}    \\
                                                                    & FGVP (Mask)                  & 69.30 \G{+9.9}    					& 45.95 \R{-4.6}      					& 43.88 \R{$<$-5} 							& 61.00 \R{$<$-5}   			                      & 65.0 \R{$<-\ $5}              & 59.2 \R{$<-\ $5}                     \\
                                                                    & FGVP (RBM)                   & 69.40 \G{+10.0}     				& 46.15 \R{-4.4}      					& 52.50 \G{+1.9}          					& 60.20 \R{$<$-5} 				                      & 79.6 \R{-4.7}                      & 92.5 \R{$<-\ $5}                     \\
                                                                    & SoM                          & 65.30 \G{+5.9}    					& 45.00 \R{$<$-5} 						& 48.33 \R{-2.22}         					& 58.90 \R{$<$-5} 				                      & 65.8 \R{$<-\ $5}              & 56.1 \R{$<-\ $5}                     \\
                                                                    & Ours (CLIP)                  & 69.50 \G{+10.1}   					& \textbf{51.50} \G{+0.9}        		& 50.96 \G{+0.4}         					& \textbf{67.70} \G{+0.7}                              & \textbf{85.3} \G{+1.0}                      & 103.3 \G{+1.3}    \\
                                                                    & Ours (LLaVA)                 & \textbf{71.01} \G{+11.6} 			& 50.80 \G{+0.2}        				& \textbf{51.38} \G{+0.8}         			& 67.10 \G{+0.1}          		                      & 84.7 \G{+0.3}                      & \textbf{103.6} \G{+1.6}    \\\midrule
\multirow{7}{*}{Gemini}             & w/o prompt                   & 50.28                     & 56.68                      & 35.11                     & 59.0                   & 78.6                     & 81.5      \\
                                    & $\quad+$Step-by-Step                      & 22.82 \R{$<-\ $5}              & 21.51 \R{$<-\ $5}              & 36.37 \G{+1.3}         & 30.6 \R{$<-\ $5} & 29.8 \R{$<-\ $5}              & 40.5 \R{$<-\ $5}                     \\
                                    & FGVP (Mask)                  & 52.88 \G{+2.6}          & 40.81 \R{$<-\ $5} & 34.88 \R{-0.2}         & 45.8 \R{$<-\ $5} & 71.0 \R{$<-\ $5}              & 64.2 \R{$<-\ $5}                     \\
                                    & FGVP (RBM)     & 53.01 \G{+2.7}         & 45.67 \R{$<-\ $5} & 34.08 \R{-1.0}         & 52.0 \R{$<-\ $5} & 77.4 \R{-1.2}                       & 82.3 \G{+0.8}    \\
                                    & SoM & 51.25 \G{+1.0}         & 27.29 \R{$<-\ $5} & 34.77 \R{-0.3}         & 34.4 \R{$<-\ $5} & 69.8 \R{$<-\ $5}              & 64.5 \R{$<-\ $5}                     \\
                                    & Ours (CLIP)                  & \textbf{58.58} \G{+8.3}         & \textbf{59.07} \G{+2.4}         & 37.71 \G{+2.6}         & \textbf{60.5} \G{+1.5}          & \textbf{80.2} \G{+1.6}                       & \textbf{85.2} \G{+3.7}    \\ 
                                    & Ours (LLaVA)                 & 58.17 \G{+7.9}         & 58.35 \G{+1.7}         & \textbf{38.16} \G{+3.1}         & 60.1 \G{+1.1}          & 80.0 \G{+1.4} & 82.3 \G{+0.8}        \\ \bottomrule
\end{tabular}}
\end{adjustbox}
\end{table}

\subsection{Comprehensive VQA Tasks}
\myPara{Datasets} Experiments are conducted on 6 datasets: VisWiz~\cite{viswiz}, TextVQA~\cite{textvqa}, MMMU~\cite{mmmu}, MME~\cite{mme}, MM-Vet~\cite{mmvet}, and LLaVA-Bench~\cite{llava}. The performance on the first four datasets is evaluated using matching accuracy with the ground truth response. The performance of the latter two datasets is measured using the GPT-based evaluation scores.

\myPara{LVLMs} Experiments are conducted using two open-source models: CogVLM~\cite{cogvlm} and LLaVA~\cite{llava15}, and two commercial models: GPT-4V~\cite{gpt4v} and Gemini~\cite{gemini}. 
Due to GPT-4V's token limit, following the experiment protocol in the previous work~\cite{som} when conducting experiments with GPT-4V, for VisWiz, TextVQA, and MMMU, we randomly selected 200 images from the dataset to verify our method. Because, about 50 questions on MM-Vet are categorised as related to personal identification or brand evaluation due to GPT-4V's safety policy and are refused to answers. Therefore, we evaluated our method's performance only on the remaining questions.

\myPara{Comparison} We compare with the following methods: (1) naively feeding the query and image to the model without any prompt; (2) using ``Let’s think step by step'' as a prompt to trigger the model's chain-of-thought process, a method that has been proven to significantly improve zero-shot reasoning performance for LLMs~\cite{sbs}; and (3) two visual prompting methods designed for LVLMs, FGVP~\cite{fgvp} and SoM~\cite{som}. FGVP is designed to generate diverse visual prompts. We compared the most straightforward method of using a mask as a visual prompt and the best-performing method of using a Reverse Blur Mask (RBM) as a visual prompt. Performance improvements/decrements listed in the table are calculated relative to the ``w/o prompt'' method.

The main observations from the experimental results are as follows:
(1) Our method consistently achieves the best performance across all datasets-LVLM pairs in \cref{tab:main}. Regardless of whether the CLIP or LLaVA is used as the auxiliary model, our method leads to performance improvements. 
For LLaVA, CogVLM, GPT-4V, and Gemini, the average improvements relative to ``w/o prompt'' are 1.94\%, 1.38\%, 1.76\%, and 3.42\%, respectively. 
Our method performs particularly well on Gemini+VisWiz, with an average improvement of 8.1\%. Excluding it, our method appears more effective for open-ended questions, with an average improvement of 2.20\% on MM-Vet and LLaVA-Bench, while the average accuracy increase on multiple-choice and true-false datasets is 1.18\%.
(2) The ``let's think step by step'' approach, which is significantly effective in LLMs, does not perform well in VQA tasks. We suspect this is because this method cannot enhance the LVLM's visual perception capabilities and may even exacerbate LVLM's hallucination due to its language-oriented prompt nature.
(3) Previous visual prompting methods, lacking the ability to adapt to different queries, do not perform well on VQA tasks. Our method is clearly superior to them. This indicates that indiscriminately annotating objects in an image does not effectively assist the model in performing VQA tasks. Visual prompting methods need the ability to adapt to queries.

\subsection{Ablation Studies}

\begin{table}[tb]
  \caption{ Ablation study on the auxiliary VLM Scale. The best result are marked for each auxiliary model-dataset pair.
   }
  \label{tab:abl1}
  \centering
    \begin{tabular}{ccc}\toprule
    Mask Model     & MMMU  & MME   \\\midrule
    w/o prompt     & 35.15 & 85.50 \\\midrule
    CLIP-ViT-B     & 36.03 \G{+0.88} & 83.50 \R{-2.00} \\
    CLIP-ViT-L     & 36.21 \G{+1.09} & 83.50 \R{-2.00} \\
    CLIP-ViT-L-336 & \textbf{37.52} \G{+2.37} & \textbf{87.16} \G{+1.66} \\\midrule
    LLaVA-7B       & 35.86 \G{+0.71} & 85.66 \G{+0.16} \\
    LLaVA-13B      & \textbf{36.95} \G{+1.80} & \textbf{86.34} \G{+0.84}\\\bottomrule
    \end{tabular}
\end{table}
\begin{table}[tb]
  \caption{ Ablation study on the mean filter kernel size. 
  }
  \label{tab:abl2}
  \centering
    \begin{tabular}{ccc}\toprule
    Kernel Size                                       & MMMU & MME  \\\midrule
    \makecell{ w/o filter\vspace{-0.4em}\\ \scriptsize (kernel size = 1)}    & 36.09 \G{+0.94}     & 83.70 \R{-1.80}\\
    3                                                                       & \textbf{36.95} \G{+1.80}        & 86.20  \G{+0.70} \\
    7                                                                        & 36.32  \G{+1.17}                         & \textbf{87.14}   \G{+1.74}\\ \midrule
    \makecell{w/o prompt\vspace{-0.4em}\\ \scriptsize (kernel size$\ge2$W$=2$H)} & 35.15            & 85.50 \\ \bottomrule
    \end{tabular}
\end{table}
\begin{table}[tb]
  \caption{ Ablation study on the Transformer layer for attribution map extraction. The best result are marked for each auxiliary model-dataset pair.
  }
  \label{tab:abl3}
  \centering
\begin{tabular}{cccc}
\toprule
\multicolumn{1}{l}{Mask Model} & Layer Index & MMMU  & MME   \\ \midrule
w/o prompt     & \na & 35.15 & 85.50 \\\midrule
\multirow{4}{*}{CLIP}          & 23          & 36.32 \G{+1.17}               & \textbf{87.16} \G{+1.66} \\
                               & 22          & \textbf{37.52} \G{+2.37}      & 84.80 \R{-0.70}\\
                               & 20          & 37.12  \G{+1.97}              & 83.20 \R{-1.30}\\
                               & 15          & 36.14 \G{+0.99}           & 83.16 \R{-1.34}\\\midrule
\multirow{4}{*}{LLaVA}         & 23          & 36.15  \G{+1.00}              & 83.10 \R{-1.40}\\
                               & 22          & 36.49  \G{+1.35}              & 83.00 \R{-1.50}\\
                               & 20          & \textbf{36.95} \G{+1.80}  & \textbf{86.34} \G{+0.84}\\
                               & 15          & 36.32  \G{+1.17}          & 83.16 \R{-2.34} \\ \bottomrule 
\end{tabular}
\end{table}

We identify three important factors affecting the performance of our method and conduct ablation studies on them.

\myPara{The Power of the Auxiliary Model} On the MMMU and MME datasets, we used CLIP models and LLaVA models of different scales to generate heatmaps, with LLaVA serving as the inference model, to compare performance. The results are shown in \cref{tab:abl1}. As the scale of the auxiliary model increased, the performance of our method also improved. Both increasing the depth of the auxiliary model or reducing the patch size to generate attribution map with finer granularity prove to be effective for improving the performance of our method. When the capability of the auxiliary model is insufficient, the masks generated by it could even be detrimental.

\myPara{The Kernel Size of the Mean Filter} To mitigate the limitations of rectangular mask patterns when highlighting irregularly shaped objects, we incorporated a mean filter into our method. We conducted ablation studies on different kernel sizes on the MMMU and MME datasets with LLaVA as the inference model. The results are shown in \cref{tab:abl2}. Without the mean filter, heatmaps with rectangular patterns could potentially harm the final task's performance. The optimal kernel size varied across datasets, due to the different image complexity and question complexity.

\myPara{The Layer for Attribution Map Extraction} Another factor affecting our method's performance was the layer used for extracting the attribution map. Although we knew that deeper layers, which contain higher-level semantic information, should be used, the specific choice of layer also impacted our method's performance. We conduct ablation on MMMU and MMe datasets using LLaVA as inference model.  The results are shown in \cref{tab:abl3}. For the CLIP model, the last two layers are more effective. However, for LLaVA, directly using the attention maps from the last two layers do not yield good results; the best performance occurred when a mid-to-late layer was used, such as $20$-th layer for LLaVA-13B.

\subsection{Self-Reflection}
\begin{table}[tb]
  \caption{ The comparison between our method and textual self-reflection method and their combination.
  }
  \label{tab:selfreflection}
  \centering
\begin{tabular}{lc}
\toprule
Prompt Method                     & LLaVA-Bench \\ \midrule
w/o prompt                        & 71.90      \\
textual self reflection             & 72.90  \textcolor{mygreen}{$(+1.00)$}    \\
ours (LLaVA)                      & 74.80   \textcolor{mygreen}{$(+2.90)$}   \\
$\quad+$ reflection via re-emphasize           & 72.70 \textcolor{mygreen}{$(+0.80)$}     \\
$\quad+$ reflection via evaluation     & \textbf{76.10}  \textcolor{mygreen}{$(+4.20)$}    \\ \bottomrule
\end{tabular}
\end{table}
When the auxiliary LVLM and the inference LVLM are the same, our method can be seen as having a two-round chat with the LVLM. The first round generates an annotated image, where the highlighted areas represent what the LVLM considers important, embedding the LVLM's process of extracting visual information. The second round conducts inference based on the generated annotated image, allowing the LVLM to perform Self-Reflection and refine its previous process of visual information extraction. Unlike previous Self-Reflection methods using text as a medium in LLMs, under the \promptname framework, all information related to the first answer is stored in the annotated image, and the text response from the first round is not provided to the model in the second round. 

As a new perspective of Self-Reflection, we explore two questions: (1) \textit{Can visual mediums also achieve effective Self-Reflection?} To answer this, we compared text-based Self-Reflection and our method using LLaVA as the inference model on the LLaVA-Bench dataset. The results in \cref{tab:selfreflection} show that our method achieves better performance than text-based Self-Reflection, proving that visual mediums can effectively facilitate Self-Reflection.

The second question is: (2) \textit{Can we more effectively utilize visual mediums for Self-Reflection?} Generally, Self-Reflection techniques involve two steps int the second round: first, evaluating the previous answer, and second, combining the evaluation to re-answer the question. However, in our framework, the evaluation process is not included, and the model directly proceeds to inference. Therefore, we designed a new inference process. We input the annotated image and the question into the VLM, prompting it to judge whether the highlighted areas in the image support the answer to the question. If yes, the answer is generated using the annotated image; if not, the answer is generated using the original image. The result (the last row of \cref{tab:selfreflection}) shows that this strategy further improves our method. Conversely, when we do not allow the model to perform evaluation and emphasize that the answer lies within the highlighted areas of the annotated image, performance decreases (second to last row in \cref{tab:selfreflection}). This also proves the importance and effectiveness of the evaluation process when using visual mediums for Self-Reflection.

\subsection{Other Discussion}
\begin{table}[tb]
  \caption{ The performance of our method on
  hallucination datasets. 
  }
  \label{tab:hallucination}
  \centering
    \begin{tabular}{l@{$\quad$}c@{$\quad$}c}
    \toprule
    Prompt Method & VisWiz-Unanswerable & POPE \\ \midrule
    w/o prompt    & 81.41               & 81.00 \\
    Ours (CLIP)   & 83.83 \textcolor{mygreen}{$(+2.42)$}              & 82.81 \textcolor{mygreen}{$(+0.81)$} \\
    Ours (LLaVA)  & \textbf{85.26} \textcolor{mygreen}{$(+3.85)$}             & \textbf{83.52} \textcolor{mygreen}{$(+2.52)$} \\ \bottomrule
    \end{tabular}
\end{table}
\myPara{Hallucination} We also explore our method's ability to assist LVLM in overcoming hallucinations. We conduct two experiments. First, on VisWiz, we calculated the accuracy with which our method and the baseline identify the unanswerable questions. These questions often involve information that does not exist in the image, thus the responses to these questions are based on hallucination. Second, we conduct experiments on a subset of a commonly used LVLM hallucination dataset POPE~\cite{Li-hallucination-2023}. The experimental results presented in \cref{tab:hallucination} demonstrate that our method also has the ability to mitigate hallucination.

\section{Conclusion}

In this work, we introduce a novel visual prompting technique called Attention Prompting on Image (\promptname), which incorporates an auxiliary LVLM to generate an attention heatmap on the image dependent on text query. Our extensive experiments demonstrate the advantages of our prompting method for different LVLMs on various benchmarks. Additionally, our approach offers new insights into using visual signals for LVLM ensembling and LVLM self-reflection. 

\section*{Acknowledgement}
This project is supported by the Ministry of Education, Singapore, under its Academic Research Fund Tier 2 (Award Number: MOE-T2EP20122-0006).

\clearpage  

%
%
\bibliographystyle{splncs04}
\bibliography{main}

\clearpage
\setcounter{figure}{1}
\setcounter{table}{6}
\begin{center}
    {\Large
    \textbf{Attention Prompting on Image for Large Vision-Language Models\\[0.5em]
    \textit{- Supplementary Material -}}}
\end{center}

\section{Examples}
\begin{figure}[h]
  \centering
  \includegraphics[width=\textwidth]{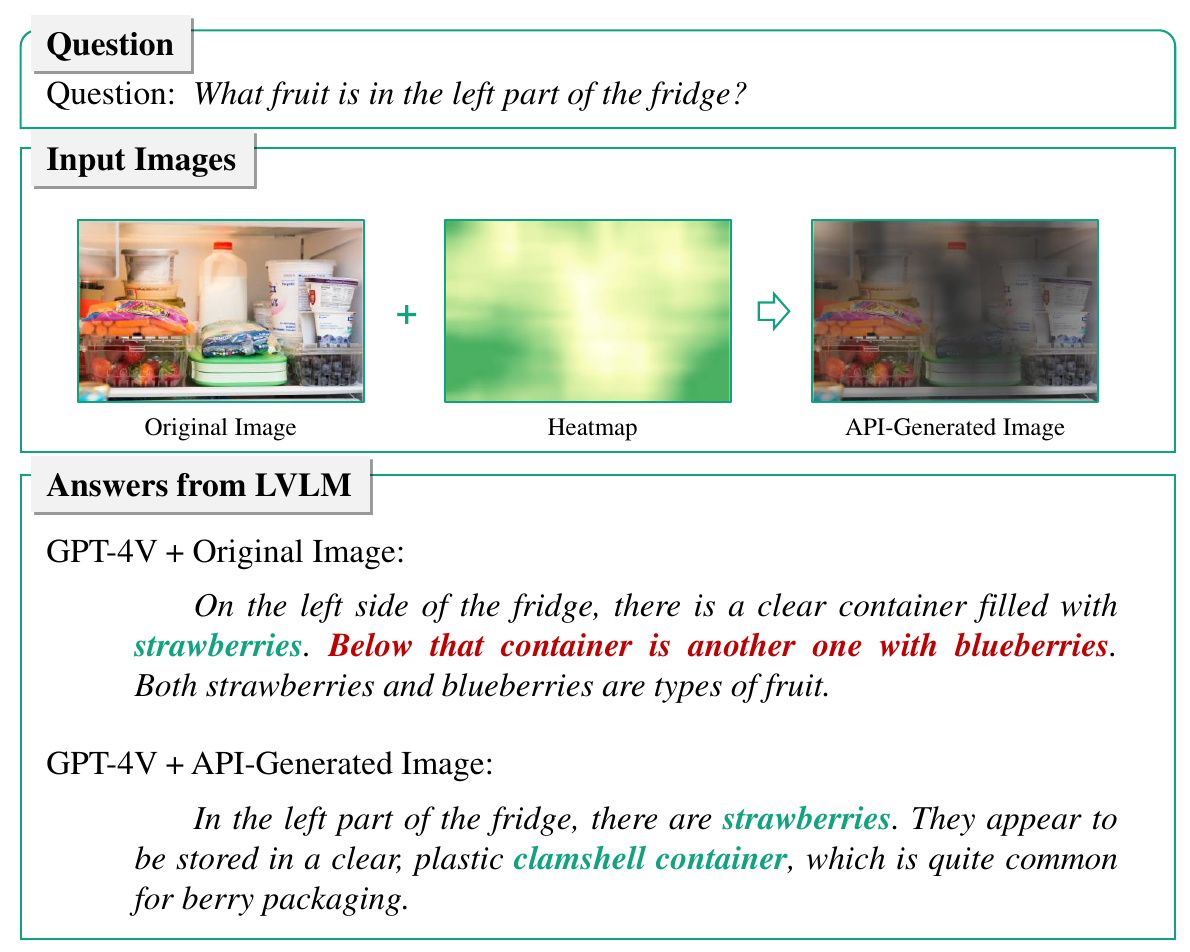}
  \caption{In complex images including multiple objects, our method accurately highlights the fruits and masks the other objects, thereby simplifying the scene and facilitating the LVLM's inference of spatial relationships.}
  \label{fig:case1}
\end{figure}
\textcolor{white}{empty}
\vfill
\clearpage

\begin{figure}[t]
  \centering
  \includegraphics[width=\textwidth]{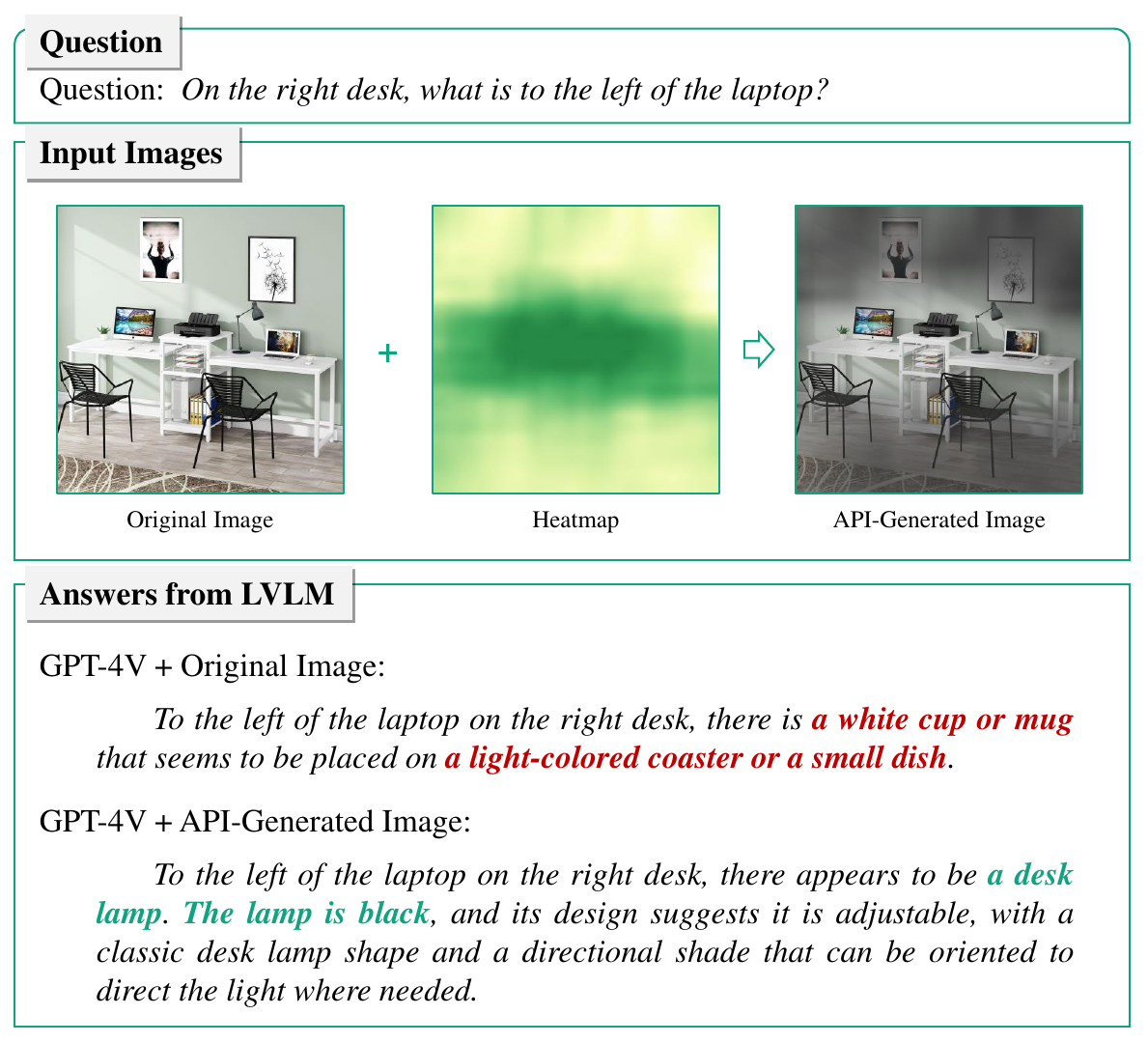}
  \caption{Our method identifies regions related to the objects, thereby assisting the LVLM in spatial reasoning.}
  \label{fig:case2}
\end{figure}
\textcolor{white}{empty}
\vfill
\clearpage



\begin{figure}[tb]
  \centering
  \includegraphics[width=\textwidth]{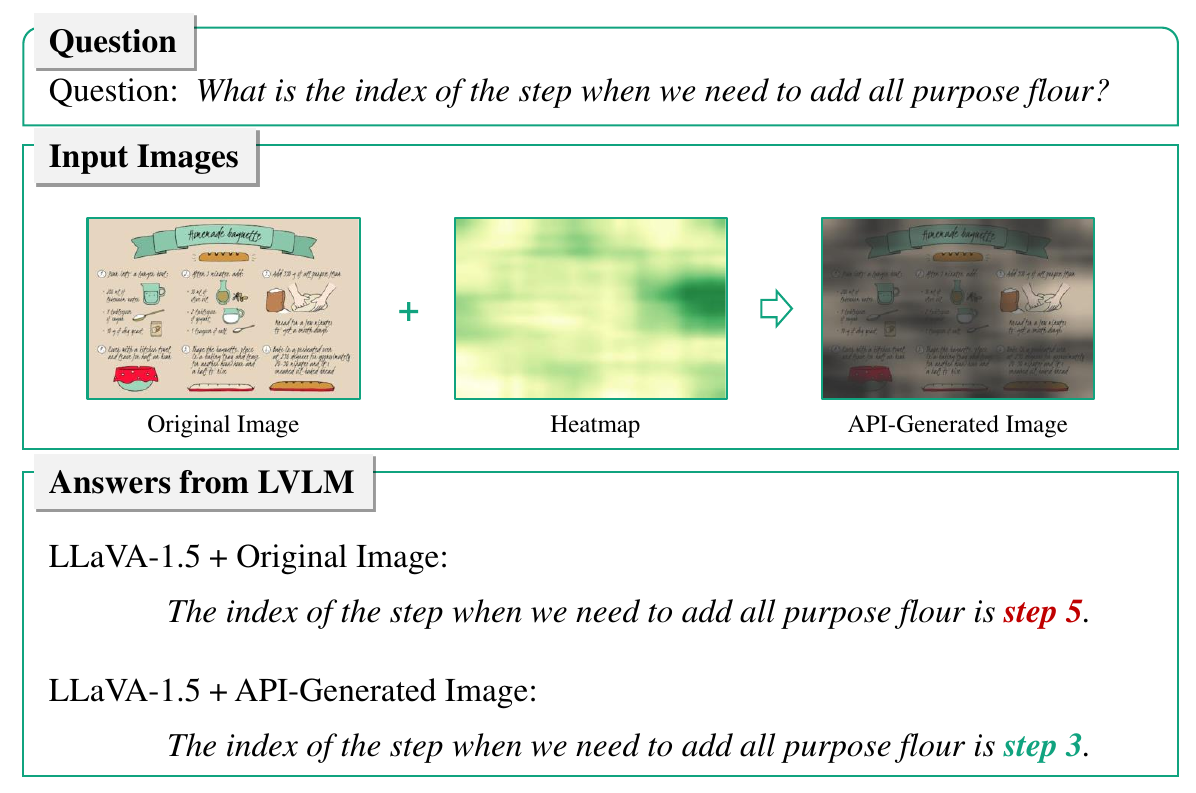}
  \caption{Our method assists LVLM's recognition process by highlighting the corresponding steps in the flowchart.}
  \label{fig:case3}
\end{figure}
\vfill
\begin{figure}[tb]
  \centering
  \includegraphics[width=\textwidth]{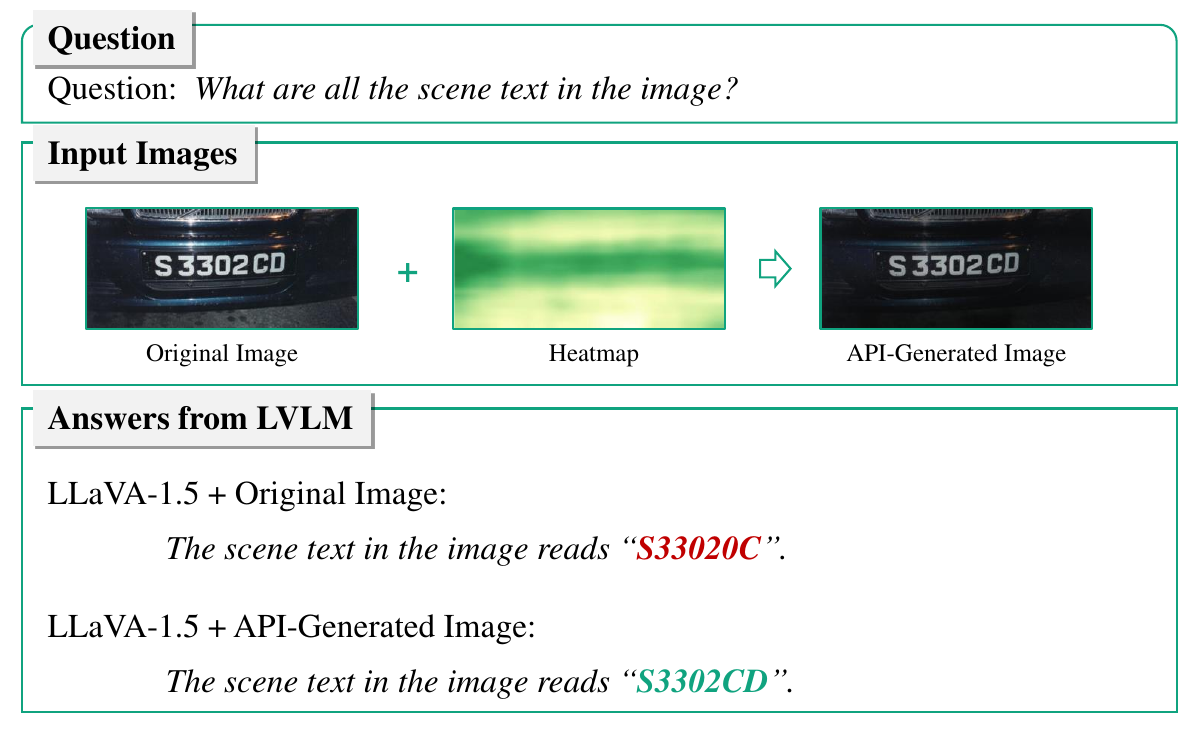}
  \caption{In this example, our method enhances LVLM's OCR capability by masking background areas and highlighting the regions that require OCR.}
  \label{fig:case4}
\end{figure}
\vfill
\clearpage

\begin{figure}[t]
  \centering
  \includegraphics[width=\textwidth]{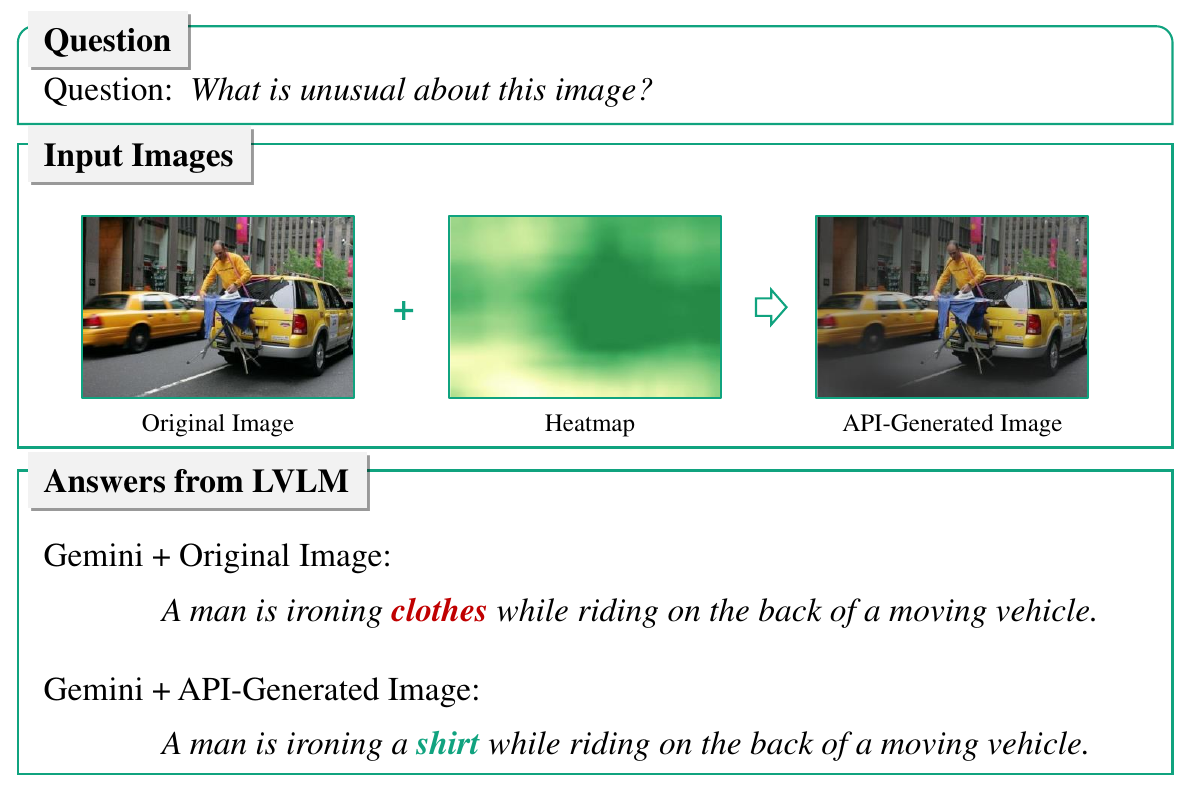}
  \caption{In this example, our method highlights related regions and enables the LVLM to generate more detailed and accurate response.}
  \label{fig:case5}
\end{figure}
\textcolor{white}{empty}
\vfill
\clearpage

\begin{figure}[t]
  \centering
  \includegraphics[width=\textwidth]{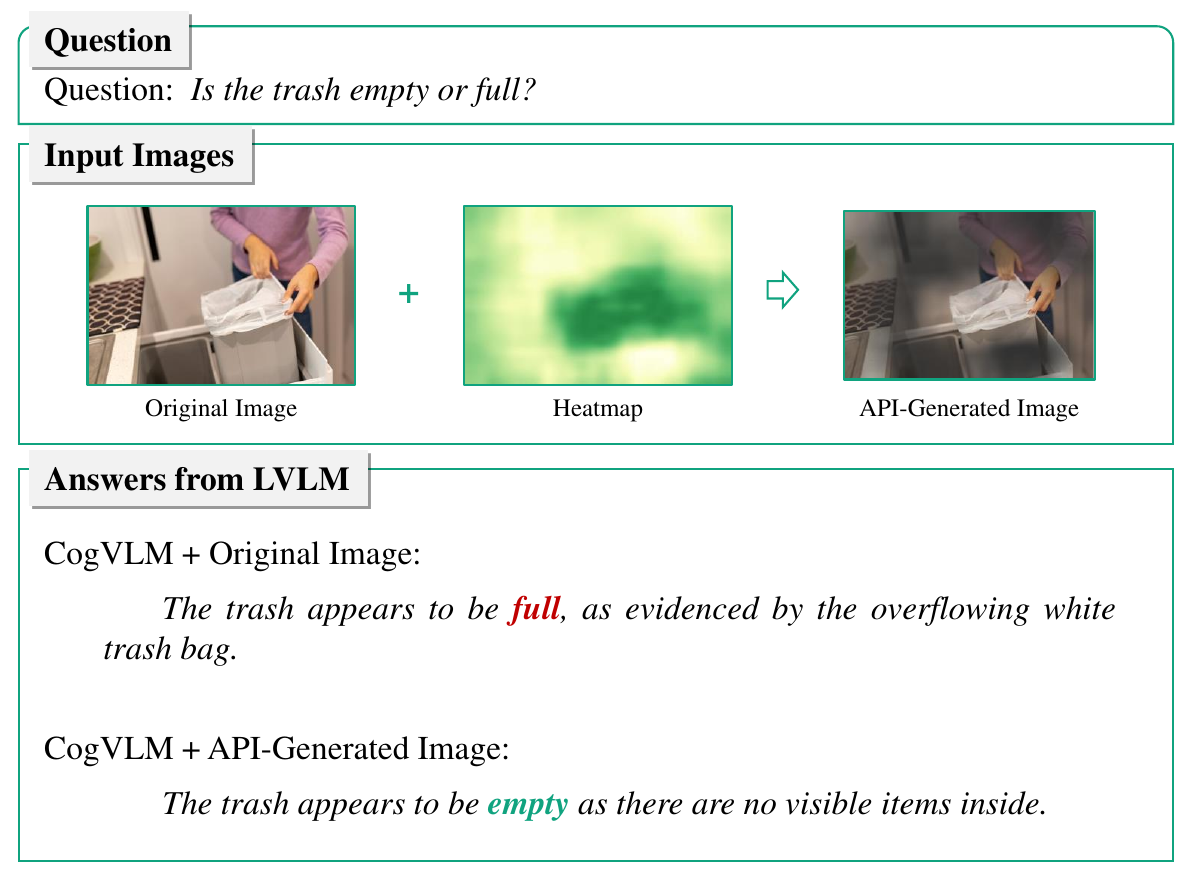}
  \caption{In this example, where the question asks to determine whether the trash can is full, our method accurately highlights the area around the trash can's opening, thereby guiding the LVLM to make a correct judgment.}
  \label{fig:case6}
\end{figure}
\textcolor{white}{empty}
\vfill
\clearpage

\begin{figure}[tb]
  \centering
  \includegraphics[width=\textwidth]{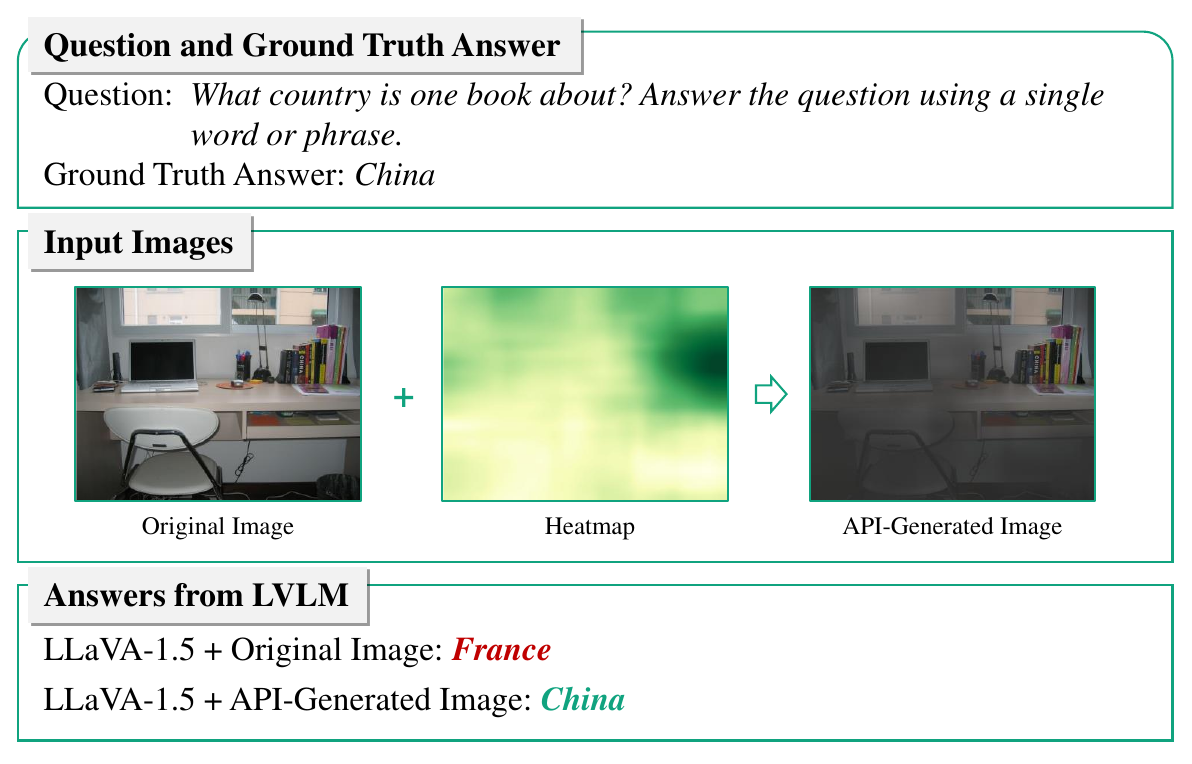}
  \caption{In this example, where the question is related to books, our method accurately highlights the area where the books are located in the image.}
  \label{fig:case7}
\end{figure}
\textcolor{white}{empty}
\vfill
\clearpage

\begin{figure}[tb]
  \centering
  \includegraphics[width=\textwidth]{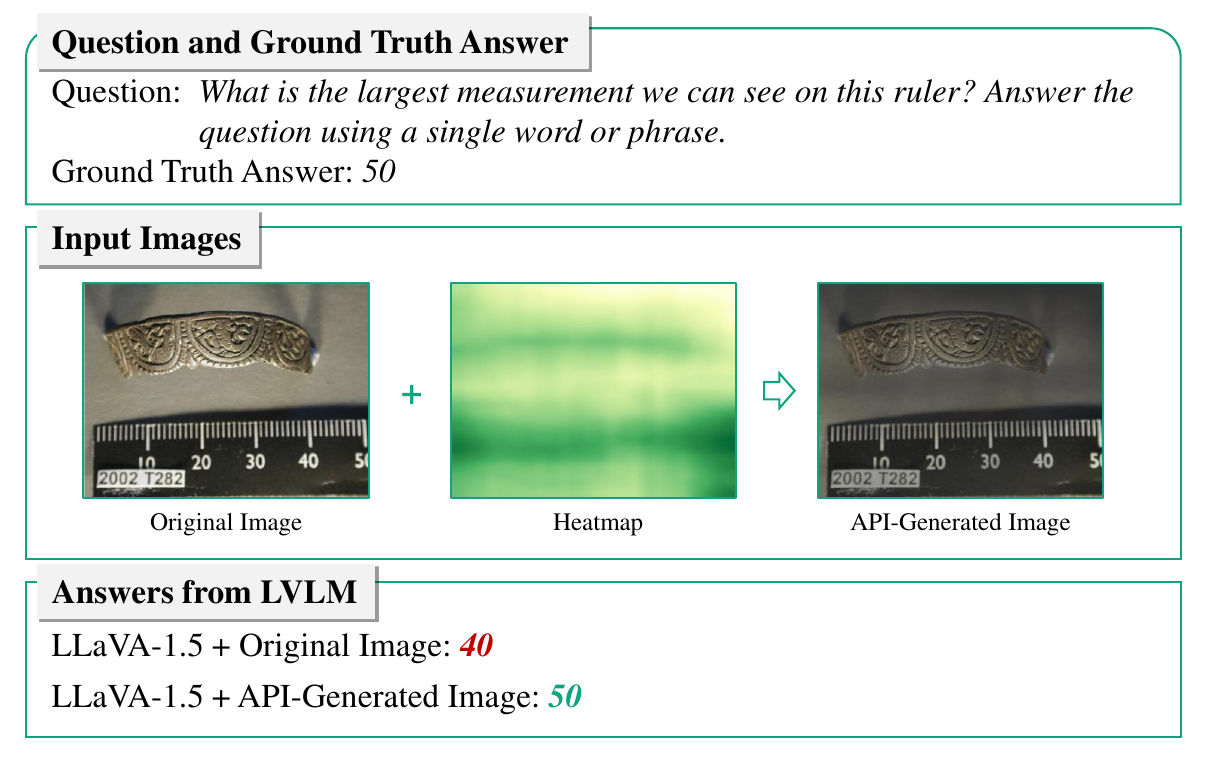}
  \caption{In this example, the largest measurement number 50 on the ruler is not fully displayed, leading to error in the baseline method. In contrast, as seen through the heatmap, our method emphasizes the bottom right corner of the image where the end of the ruler is located, thereby guiding the LVLM to provide the correct answer.}
  \label{fig:case8}
\end{figure}
\textcolor{white}{empty}
\vfill
\clearpage

\begin{figure}[tb]
  \centering
  \includegraphics[width=\textwidth]{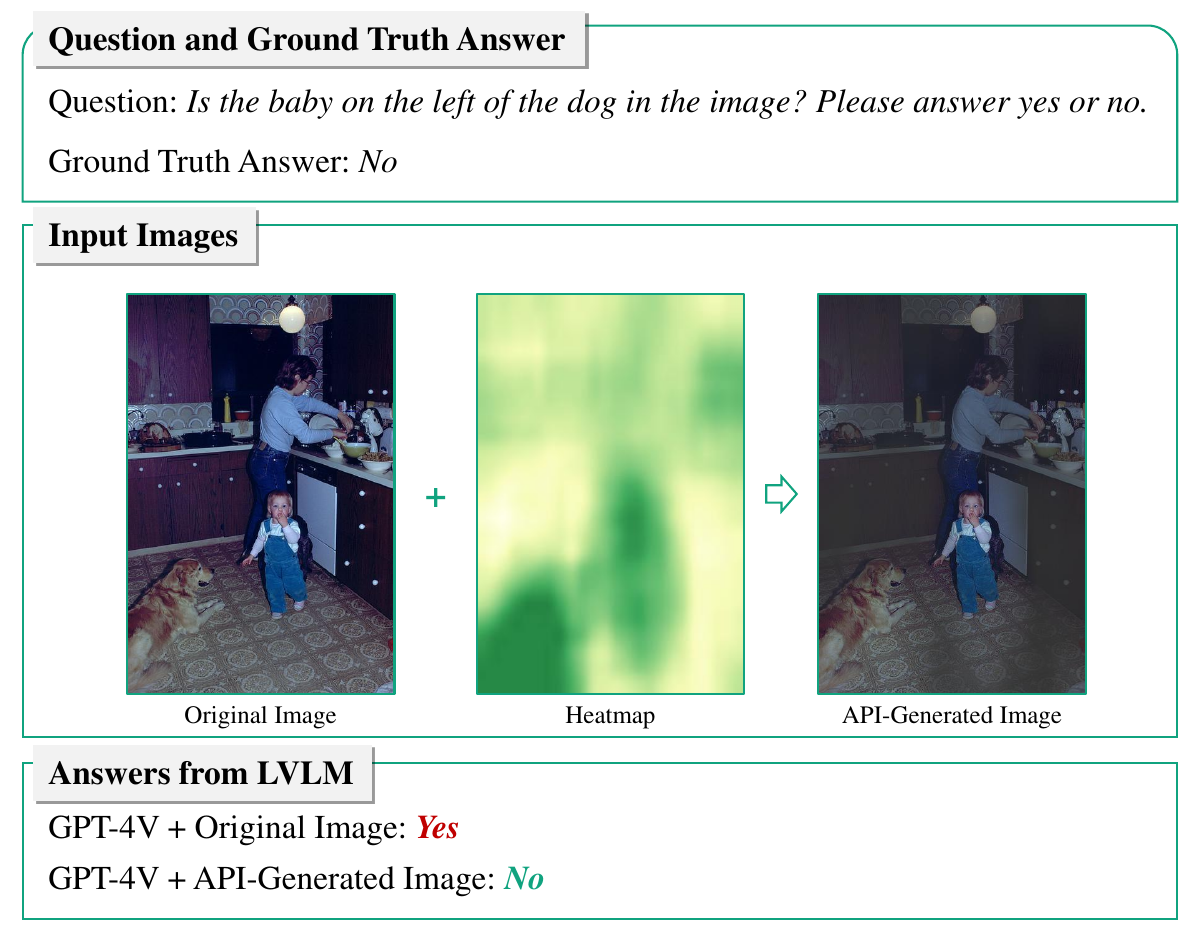}
  \caption{Our method accurately emphasizes the baby and dog in the image, thereby facilitating the inference of their spatial relationship.}
  \label{fig:case9}
\end{figure}
\textcolor{white}{empty}
\vfill
\clearpage

\begin{figure}[tb]
  \centering
  \includegraphics[width=\textwidth]{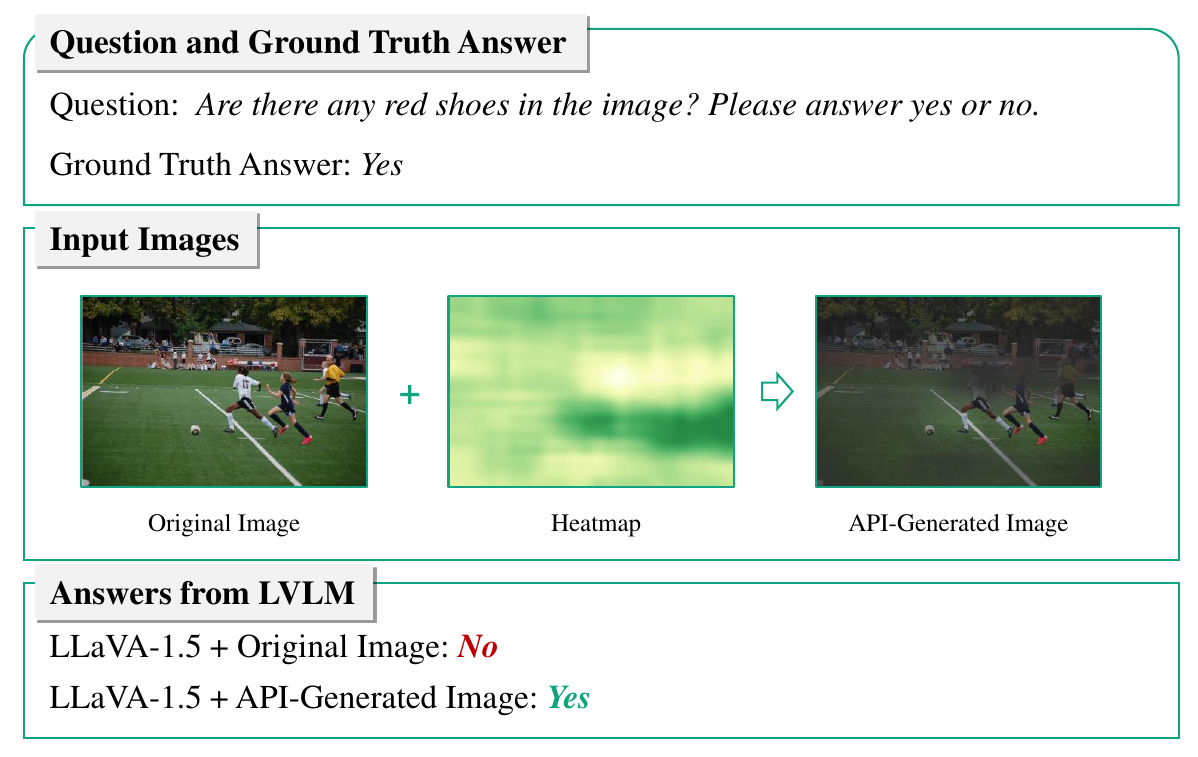}
  \caption{In this example, the question is related to the shoes, which are small objects and are difficult to recognize for the model. Our method accurately located the shoes in the image, leading the LVLM to the correct answer.}
  \label{fig:case10}
\end{figure}

\section{Notation Table}
Although the definitions of all symbols are included within the main text, we provide a comprehensive notation table in \cref{tab:notation_1,tab:notation_2} to facilitate easy reference and a macro-level understanding of the concepts involved in each part of the method.

\begin{table}[h]
  \caption{The notations used in the manuscript.
  }
  \label{tab:notation_1}
  \centering
\begin{tabular}{@{}cp{8cm}c@{}}
\toprule
\textbf{Symbol}                     & \makecell{\centering \textbf{Definition}}                                                                                                                                                                                                                                                                                                                                                                                                                                        & \textbf{Mainly used in} \\ \midrule
$f$                        & LVLM used for inference                                                                                                                                                                                                                                                                                                                                                                                                                              & Entire Sec. 3  \\ \midrule
$g$                        & Auxiliary LVLM used for   attribution map extraction                                                                                                                                                                                                                                                                                                                                                                                                 & Entire Sec. 3  \\ \midrule
$\mathcal{A}$              & Annotation function, which is   the proposed method                                                                                                                                                                                                                                                                                                                                                                                              & Entire Sec. 3  \\ \midrule
$I$                        & Original image                                                                                                                                                                                                                                                                                                                                                                                                                                       & Entire Sec. 3  \\ \midrule
$I^a$                      & Image with annotations, which is   obained by visual prompting method                                                                                                                                                                                                                                                                                                                                                                                & Entire Sec. 3  \\ \midrule
$\Psi$                     & Attribution map in the token   space, which is extracted from the auxiliary LVLM and is used to generate the   heatmap                                                                                                                                                                                                                                                                                                                               & Entire Sec. 3  \\ \midrule
$\Phi$                     & Heatmap in the pixel space,   which will be overlied on the original image                                                                                                                                                                                                                                                                                                                                                                           & Entire Sec. 3  \\ \midrule
$T^i$                      & Input text query                                                                                                                                                                                                                                                                                                                                                                                                                                     & Entire Sec. 3  \\ \midrule
$T^o$                      & Output text response                                                                                                                                                                                                                                                                                                                                                                                                                                 & Entire Sec. 3  \\ \midrule
$A^{(l,h)}$                & Attention map in the $l$-th transformer  layer corresponding to the $h$-th head                                                                                                                                                                                                                                                                                                                                                                                 & Entire Sec. 3  \\ \midrule
$g_{\text{clip}}$          & CLIP model                                                                                                                                                                                                                                                                                                                                                                                                                                           & Sec. 3.1       \\ \midrule
$\hat{I}$                  & Image feature generated by CLIP,   which is able to calculate the similarity                                                                                                                                                                                                                                                                                                                                                                         & Sec. 3.1       \\ \midrule
$\hat{T}$                  & Text feature generated by CLIP,   which is able to calculate the similarity                                                                                                                                                                                                                                                                                                                                                                          & Sec. 3.1       \\ \midrule
$L$                        & Number of transformer layers   within the CLIP vision encoder                                                                                                                                                                                                                                                                                                                                                                                        & Sec. 3.1       \\ \midrule
MSA                        & Multihead Self-Attention   structure                                                                                                                                                                                                                                                                                                                                                                                                                 & Sec. 3.1       \\ \midrule
MLP                        & Multi-Layer Perceptron structure                                                                                                                                                                                                                                                                                                                                                                                                                     & Sec. 3.1       \\ \midrule
$Z^l$                      & Input token sequence for the   $l$-th transformer layer                                                                                                                                                                                                                                                                                                                                                                                              & Sec. 3.1       \\ \midrule
$[Z]_{\text{cls}}$         & Value of the cls token within   the token sequence $Z$.                                                                                                                                                                                                                                                                                                                                                                                              & Sec. 3.1       \\ \bottomrule
\end{tabular}
\end{table}

\begin{table}[tb]
  \caption{The notations used in the manuscript.
  }
  \label{tab:notation_2}
  \centering
\begin{tabular}{@{}cp{8cm}c@{}}
\toprule
\textbf{Symbol}                     & \makecell{\centering \textbf{Definition}}                                                                                                                                                                                                                                                                                                                                                                                                                                        & \textbf{Mainly used in} \\ \midrule
$\mathcal{L}$              & Linear transformation in the   CLIP model, which is performed after the transformer structure, before   calculating the similarity score                                                                                                                                                                                                                                                                                                             & Sec. 3.1       \\ \midrule
$L'$                       & In the similarity decomposition   of the CLIP model, only the MSA output of last $L-L'$ layers are considered.   $L'$ is the starting layer index.                                                                                                                                                                                                                                                                                                   & Sec. 3.1       \\ \midrule
$V^{(l,h)}$                & Value matrix in the $l$-th layer   corresponding to the $h$-th head                                                                                                                                                                                                                                                                                                                                                                                  & Sec. 3.1       \\ \midrule

$W^{(l,h)}$                & Weight matrix in the $l$-th   layer used to merge the multiple attention heads and corresponds to the   $h$-th head. For each head, after the the multiplication between the   attention map and the value matrix, we have a matrix with the size of   $T\times D'$. To aggregate the matrices from all heads, a weight matrix with   the size of $(H\times D')\times D$ is used. $W^{(l,h)}$ is obtained from   splitting this large weight matrix. & Sec. 3.1       \\ \midrule
$B^{(l)}$                  & Bias matrix in the $l$-th layer   used to merge the multiple attention heads                                                                                                                                                                                                                                                                                                                                                                         & Sec. 3.1       \\ \midrule
$A^{(l,h)}_{\text{cls},t}$ & Attention value of the class   token towards the $t$-th token in $A^{(l,h)}$                                                                                                                                                                                                                                                                                                                                                                         & Sec. 3.1       \\ \midrule
$V^{(l,h)}_{t,:}$          & $t$-th row of $V^{(l,h)}$                                                                                                                                                                                                                                                                                                                                                                                                                            & Sec. 3.1       \\ \midrule
$H$                        & Number of attention heads                                                                                                                                                                                                                                                                                                                                                                                                                            & Sec. 3.1       \\ \midrule
$T$                        & Number of tokens                                                                                                                                                                                                                                                                                                                                                                                                                                     & Sec. 3.1       \\ \midrule
$\eta^l_t$                 & MSA output of the $l$-th layer   corresponding to  the $t$-th   patch(token)\}                                                                                                                                                                                                                                                                                                                                                                       & Sec. 3.1       \\ \midrule
$\psi_t$                   & $\eta^l_t$ summing over the   layer index                                                                                                                                                                                                                                                                                                                                                                                                            & Sec. 3.1       \\ \midrule
$\Psi^{cls}$               & Attribution map generated from   the CLS token                                                                                                                                                                                                                                                                                                                                                                                                       & Sec. 3.1       \\ \midrule
$\Psi^{comp}$              & Complementary attribution map   generated using the non-CLS token                                                                                                                                                                                                                                                                                                                                                                                    & Sec. 3.1       \\ \midrule
$Z^{\text{text}}$          & $N$ tokens corresponding to the   text query                                                                                                                                                                                                                                                                                                                                                                                                         & Sec. 3.2       \\ \midrule
$Z^{\text{img}}$           & $P\times P$ tokens corresponding   to the image patches                                                                                                                                                                                                                                                                                                                                                                                              & Sec. 3.2       \\ \midrule
$Z^{\text{out}}$           & $M$ tokens generated by the   LLaVA model                                                                                                                                                                                                                                                                                                                                                                                                            & Sec. 3.2       \\ \midrule
$A^{(\bar{L},h)}_{m,t}$    & Attention value in   $A^{(\bar{L},h)}$ from the $m$-th token to the $t$-th token                                                                                                                                                                                                                                                                                                                                                                     & Sec. 3.2       \\ \midrule
$\hat{\Phi}$               & Raw heatmap, which is generated   by resizing the attribution map                                                                                                                                                                                                                                                                                                                                                                                    & Sec. 3.3       \\ \bottomrule
\end{tabular}
\end{table}
\clearpage

\section{Observation and Discussion of API Method}

\subsection{CLS Token Similarity and Non-CLS Token Similarity}
\begin{figure}[tb]
  \centering
  \includegraphics[width=\textwidth]{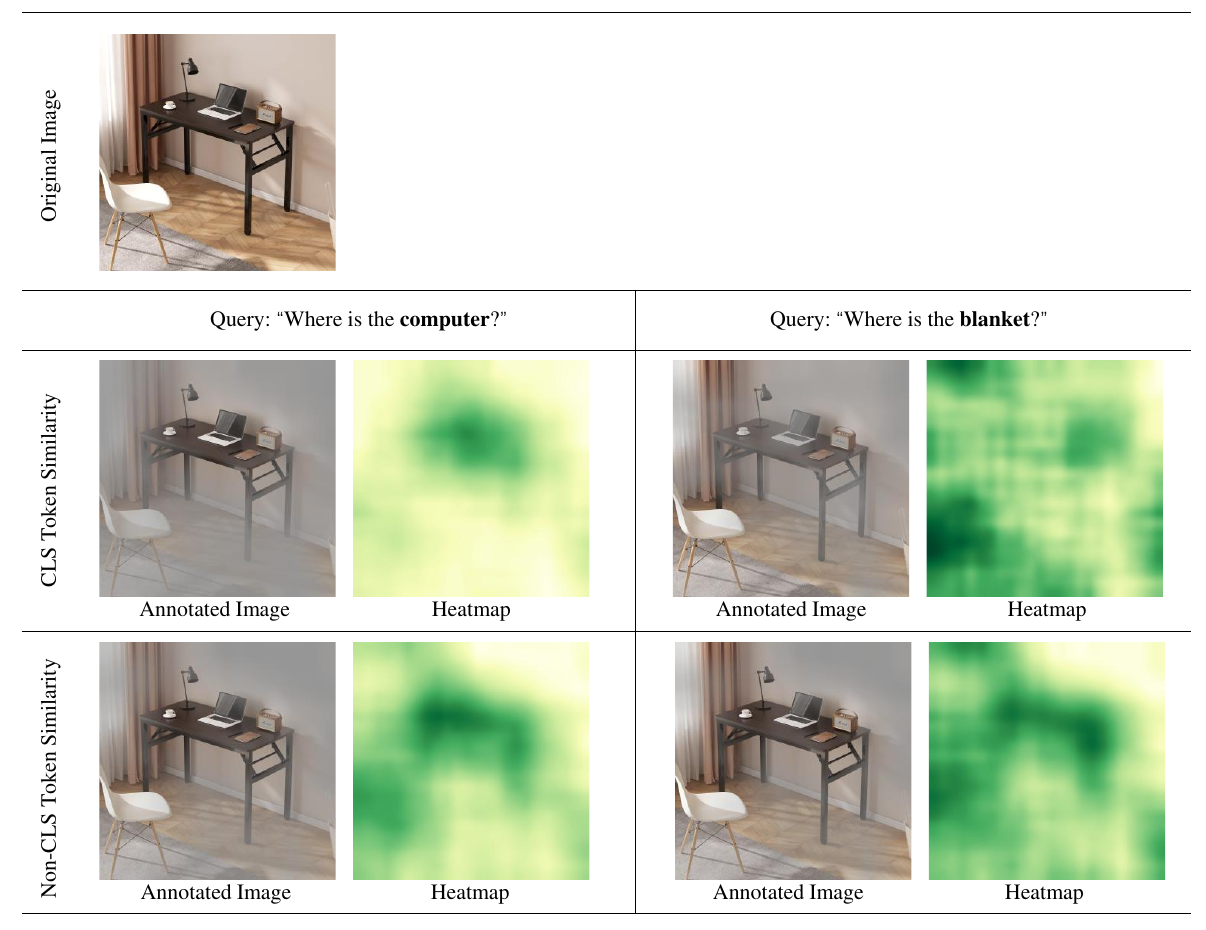}
  \caption{Comparison between the functionality of CLS token similarity and the Non-CLS token similarity.}
  \label{fig:cls}
\end{figure}
To extract heatmaps from the CLIP model, we designed two complementary types of attribution maps: one based on the decomposition of similarity between the feature of the CLS token and text feature, and the other measuring the similarity between the feature of the Non-CLS tokens and text feature. \cref{fig:cls} compares the differences in functionality between these two types of attribution maps. The third row in the image shows the heatmap generated solely based on $\Psi^{cls}$ and its resulting annotated image. The fourth row shows the heatmap obtained solely from $\Psi^{comp}$. Firstly, we can observe that when the query changes, $\Psi^{cls}$ can highlight different parts of the image corresponding to different queries. It selects the areas where the blanket and computer are located based on the query. However, $\Psi^{comp}$ does not show significant differences in response patterns to different queries. On the other hand, $\Psi^{comp}$ can filter out the background of the image, leaving the objects, which potentially can be used in the process of VQA. For instance, when the query explicitly mentions ``computer'', $\Psi^{cls}$ completely ignores the chair and blanket in the lower left corner, but $\Psi^{comp}$ still assigns high values to these areas. Therefore, we combine $\Psi^{cls}$ and $\Psi^{comp}$ to form a complete attribution map.

\subsection{Attribution Map Aggregation for CLIP Model}
First, Eq. (7) in the maintext can be rewritten as $1 - (1-\Psi^{cls})(1-\Psi^{comp})$, where since $\Psi^{cls}$ and $\Psi^{comp}$ are cosine similarities, both $(1-\Psi^{cls})$ and $(1-\Psi^{comp})$ range between 0 and 1. Thus, the final mask is related to the product of the two parts, $(1-\Psi^{cls})$ and $(1-\Psi^{comp})$. If $\Psi^{cls}$ and $\Psi^{comp}$ are considered binary, then $(1-\Psi^{cls})(1-\Psi^{comp})$ can be approximated as an OR operation between $(1-\Psi^{cls})$ and $(1-\Psi^{comp})$. That is, when either $(1-\Psi^{cls})$ or $(1-\Psi^{comp})$ is 0, the equation will be 1, and only when both are 1, the equation will be 0. This means that for patch $i$, as long as either attribution map $\Psi^{cls}$ or $\Psi^{comp}$ highlights this patch, the final attribution map $\Psi$ will also highlight this patch. Only when both $\Psi^{cls}$ and $\Psi^{comp}$ consider patch $i$ unimportant, the final attribution map will ignore this patch.

Experimental findings, as shown in \cref{fig:cls}, indicate that, on one hand, $\Psi^{comp}$ can indiscriminately choose all entities, whereas $\Psi^{cls}$ selects entities explicitly mentioned in the query. The highlighted area in $\Psi^{cls}$ can be understood as a subset of the highlighted area in $\Psi^{comp}$. On the other hand, both $\Psi^{cls}$ and $\Psi^{comp}$ will ignore non-informative parts of the image. Therefore, in actual non-binary cases, the computation of Eq. (7) can be described as an algorithm: first, apply a mask to non-informative areas (\textit{i.e.}, instruct the LVLM to ignore these patches) because these patches will not be selected by either $\Psi^{cls}$ or $\Psi^{comp}$. For the remaining areas, which are patches with objects directly mentioned in the query or other entities potentially related to the query, a multiplication of $\Psi^{cls}$ and $\Psi^{comp}$ further highlights the patches with objects appearing in the query because they have greater weight in $\Psi^{cls}$.

\section{More Experimental Results and Implementation Details}
\subsection{Ensemble}
\begin{table}[]
\centering
	\caption{Ensemble of visual prompts generated from different LVLM.
	}
	\label{tab:ensemble}
\begin{tabular}{@{}ll@{}}
\toprule
                  & LLaVA-Bench \\ \midrule
w/o prompt        & 102.00      \\
Ours (CLIP)       & 103.30 \G{+1.30}      \\
Ours (LLaVA)      & 103.60 \G{+1.60}     \\
Ours (CLIP+LLaVA) & 104.80 \G{+2.80}     \\ \bottomrule
\end{tabular}
\end{table}
When the auxiliary LVLM and the LVLM used for inference are different, our approach can be seen as ensembling the knowledge of the auxiliary LVLM into the LVLM used for inference through visual prompts. Under this definition, baseline methods like FGVP and SoM can also be considered a form of ensemble, not between LVLMs but between a vision model (segmentation model) and an LVLM. From the experimental results, our method is the first effective ensemble method that is based on visual prompting in a VQA context.

In traditional ensemble methods that are based on output aggregation, the number of models to be ensembled can be more than 2. However, in our method, we ensemble only two models, namely, an auxiliary LVLM and an LVLM for inference. To achieve an ensemble of more than two models, we conduct the following experiment. We use GPT-4V as the inference model and experiment on the LLaVA-Bench (in-the-wild) dataset, Instead of using a single annotated image. We input the annotated images generated by both \promptname+CLIP and CLIP+LLaVA simultaneously into GPT-4V, while keep using the original question without additional prompts as the textual query. The experimental results in \cref{tab:ensemble}, show that the ensemble of \promptname+CLIP and CLIP+LLaVA can further improve performance.

\subsection{Influence on Different VQA Abilities}
To thoroughly understand the impact of our method on various capabilities of LVLMs, we report the performance changes across different specific abilities on the MM-Vet dataset using the CogVLM model as the inference model and CLIP as the mask model. The results are shown in \cref{tab:percate}. It is observed that our method enhances all categories of capabilities in the MM-Vet dataset. Notably, our method is particularly beneficial for OCR and Math abilities. The significant improvement in OCR capability is attributed to our method's highlighting of relevant areas, allowing the model to focus only on regions related to answering the question. This narrows down the scope of the OCR task, thereby enhancing OCR performance. Consequently, the improvement in mathematical ability is closely linked to the enhancement in OCR capability. Since addressing math-related questions in images first requires performing OCR tasks, the improvement in OCR also contributes to the enhancement of mathematical abilities.
\begin{table}[]
  \caption{ The influence of our method on various categories of LVLM capabilities.
  }
  \label{tab:percate}
  \centering
\begin{tabular}{@{}c@{$\quad$}c@{$\quad$}c@{$ \quad$}c@{$\quad $}c@{$\quad $}c@{$\quad $}c@{}}
\toprule
\multirow{2}{*}{} & \multicolumn{6}{c}{Capability}                                            \\ \cmidrule(l){2-7} 
                        & Recognition & OCR  & Knowledge & Generation & \makecell{Spatial\\Relationship} & Math \\ \midrule
w/o prompt              & 54.9        & 42   & 43.9      & 42.6       & 50.1                 & 3.5  \\
Ours                    & 55.3        & 48.3 & 45.6      & 46         & 51.2                 & 14.6 \\ \bottomrule
\end{tabular}
\end{table}

\subsection{Implementation Details}
\myPara{Pre-trained weight and \promptname}
During the mask generation phase, we used the CLIP-ViT-L-336 model~\cite{clip} released by OpenAI and the LLaVA-1.5-13B model~\cite{llava15}. In the inference process, we utilized the released weight of LLaVA-1.5-13B model~\cite{llava15} and cogvlm-chat-v1.1 model~\cite{cogvlm}. We use the ``gpt-4-1106-vision-preview'' and ``gemini-pro-vision'' models for GPT-4V~\cite{gpt4v} and Gemini~\cite{gemini} \promptname, respectively. All local experiments were deployed on a single A100 GPU.

\myPara{Query GPT-4V and Gemini}
For GPT-4V and Gemini, we used python APIs for batch querying. When encountering errors due to server or network issues, we paused for a while and retried the query once. If the error persisted, we recorded the response as an empty string. If a query was detected against security policy, such as person identification, we did not retry and directly recorded the responses from GPT-4V and Gemini as empty strings.

\myPara{Baselines} The ``w/o prompt'’ baseline is implemented by directly querying the LVLM with the question together with the original image. Following \cite{sbs}, the ``Step-by-Step'' baseline is implemented by inputting the original image and query in the format of 
\begin{quote}
    [Question] Let's think step by step.
\end{quote}
For the experiments with FGVP~\cite{fgvp} and SoM~\cite{som}, we query the LVLM with the corresponding annotated image and the original question, which is also the same when we implement our method. The only difference among the experiments with FGVP, SoM and our method is the annotated image. For the FGVP method, the annotation process is aligned with the default of the released code. For the SoM method, we choose SAM~\cite{sam} as the segmentation model and keep all other parameters aligned with the default setting in the released code.

\myPara{Implementation on each dataset} Our implementation on various datasets adopts the approach from LLaVA~\cite{llava}. The evaluation process of each dataset adheres to its official usage protocols or its official  template, when it is accessible. 
(1) \textbf{LLaVA-Bench (in-the-Wild)}~\cite{llava} is a dataset comprising real-world scenes, drawings, memes, and other types of images, along with open-ended questions. It focuses on testing LVLMs' capabilities in QA, detailed description, and complex reasoning. In our implementation, the textual prompt is directly the question from the dataset. We record the LVLM's complete answer and use the GPT-based evaluation tool officially released by LLaVA-Bench (in-the-Wild) to score the answers. 
(2) \textbf{MM-Vet}~\cite{mmvet} is a comprehensive dataset containing various types of images, including real-world scenes, artworks, statistical graphs, memes, etc., along with open-ended questions. Each question involves multiple aspects of visual and language abilities, such as recognition + spatial awareness or OCR + Math. In our implementation, the textual prompt is directly the question from the dataset. We record the LVLM's complete answer and use the GPT-based evaluation tool officially released by MM-VET to score the answers.
(3) \textbf{MME}~\cite{mme} is a dataset that includes images of real-world scenes, artworks, logos, etc., along with True-False questions. This dataset involves abilities in commonsense reasoning, numerical calculation, and text translation, among others. Given its binary response format (yes or no), we add ``Please answer yes or no'' as an additional textual prompt to the original question. We evaluate the performance by the matching accuracy between LVLM's answers and the ground truth.
(4) The \textbf{MMMU}~\cite{mmmu} dataset encompasses multi-discipline questions requiring college-level expertise for responses. The questions are either multiple-choice or can be answered with simple data or phrases. For multiple-choice questions, we guide the LVLM to directly answer the corresponding option by adding ``Answer with the option's letter from the given choices directly'' after the original question and options. For other questions, we add ``Answer the question using a single word or phrase.'' to the original question. Our experiment is conducted using the validation set of MMMU. Evaluation is based on the matching accuracy between LVLM's answers and the ground truth. 
(5) The \textbf{TextVQA}~\cite{textvqa} dataset contains real-world images with text, where the questions can be answered with simple words or phrases, mainly testing the LVLM's OCR and reasoning abilities. We add ``Answer the question using a single word or phrase'' after the original question to guide the LVLM to directly respond to the query without providing additional explanations.  Our experiment is conducted using the validation set of TextVQA. The evaluation score is the matching accuracy between LVLM's answers and the ground truth. 
(6) The \textbf{VisWiz}~\cite{viswiz} dataset is collected from questions about real-world images asked by blind people and manually annotated answers. The questions can be answered with simple words or phrases. However, since the questions are from blind individuals, some questions are unanswerable based on the image alone and thus are marked as unanswerable. To address this, we concatenate the following prompt after the original question: ``When the provided information is insufficient, respond with 'Unanswerable'. Answer the question using a single word or phrase''  Our experiment is conducted using the validation set of VisWiz. Evaluation is based on the matching accuracy between LVLM's answers and the ground truth.

\myPara{Prompts used in the Self-Reflection experiment}
For the textual self-reflection experiment, we use a two-round chat. In the first round, we directly ask the LVLM to answer the query and record the answer. In the second round, we use a prompt in the format of 
\begin{quote}
    For the Question ``\text{[Question]}'', Your previous answer is ``\text{[Answer in the Round 1]}''. Evaluate the quality of the answer and provide a new answer.
\end{quote}
We record the response of the second round and extract the answer by manually delete the sentences related to the quality evaluation of previous answer. The extracted answer is stored as the final answer. For the ``\promptname + reflection via re-emphasize'' setup, we input the annotated image together with the prompt in the format of 
\begin{quote}
    [Question] (Hint: The answer is related to the unmasked visible regions).
\end{quote} 
For the ``\promptname + reflection via evaluation'' setup, we input the annotated image together with the prompt in the format of 
\begin{quote}
    For this image, the question is ``[Question]''. Evaluate whether the unmasked visible regions of the image alone can provide an answer to the question. If they suffice to answer the question, respond with letter ``T''. If they do not support an answer to the question, reply with the letter ``F''.
\end{quote}
If the LVLM responses with ``F'', we query it again using the original image and the question, and then use the response as the final answer. If the LVLM responses with ``T'', we query it again using the annotated image and the question, and then use the response as the final answer.

\section{Limitation, Future Direction, and Potential Impact}
\myPara{Limitation and future direction} An essential component of this work is the extraction of attribution maps based on an auxiliary LVLM. The introduction of an auxiliary LVLM enhances the performance of visual prompting methods but also introduces some limitations and new research opportunities. First, generating visual prompts based on an LVLM incurs additional computational costs, either from an extra execution of the same LVLM or a forward pass through another LVLM. Note that this is a limitation, exploring ways to reduce this additional overhead, such as using lightweight LVLMs to generate visual prompts to achieve a weak-to-strong effect~\cite{wts2,wts}, is a worthwhile research direction. Secondly, our current selection of auxiliary LVLMs is not adaptive; we cannot automatically choose a more suitable auxiliary LVLM for different image-query pairs. This is another limitation of our method and a potential research direction with promise.

\myPara{Potential impact} The potential social impacts of this work mainly include two aspects. The first aspect is the potential accumulation of bias and unfairness due to the introduction of an extra LVLM. The bias and unfairness of the auxiliary LVLM may accumulate through our visual prompts into the final inference process. The other aspect is the creation of a new possibility for attacks, namely, by attacking the auxiliary LVLM to generate harmful visual prompts, thereby attacking the LVLM. Because the attack is based on the visual prompts in the pixel space, such attacks might be more covert and difficult to detect.

\end{document}